%% file: main.tex
\documentclass[sigconf]{acmart}

\renewcommand\footnotetextcopyrightpermission[1]{} %
\setcopyright{none}
\acmDOI{}
\acmISBN{}
\acmPrice{}

\input{macros}

\date{}
\title{\fontsize{18}{20}\selectfont\textsc{\sys}: A Two-Tiered Architecture for Offline LLM Inference}

\author{
\begin{tabular}{P{4cm} p{0.1cm} P{4cm}}
{\rm Pouya Hamadanian} & & {\rm Sadjad Fouladi} \\
\emph{MIT} & & \emph{Microsoft Research}
\end{tabular}
\vspace{3ex}
}

\acmConference{}
\acmYear{}
\copyrightyear{}

\renewcommand{\shorttitle}{A Two-Tiered Architecture for Offline LLM Inference}

\renewcommand{\shortauthors}{P. Hamadanian, S. Fouladi}

\newcolumntype{P}[1]{>{\centering\arraybackslash}p{#1}}
\newcolumntype{Q}[1]{>{\raggedright\arraybackslash}p{#1}}
\newcolumntype{R}[1]{>{\raggedleft\arraybackslash}p{#1}}

\begin{document}

\begin{abstract}
    \input{sections/0-abstract}

\end{abstract}

\maketitle

\input{sections/1-introduction}

\input{sections/3-mem_comp_calculus}
\input{sections/4-kv_calculus}

\input{sections/5-design}

\input{sections/6-eval}

\input{sections/8-related}

\input{sections/9-conclusion}

\label{beforerefs}
\clearpage

\bibliographystyle{ACM-Reference-Format}
\bibliography{main}

\clearpage
\appendix
\input{sections/appendix/resources}

\input{sections/appendix/calculus}
\input{sections/appendix/simulation}
\input{sections/appendix/prefill}
\input{sections/appendix/eval}

\end{document}

%% file: macros.tex
\usepackage{xurl}
\usepackage{amsmath}
\usepackage{amsthm}
\usepackage{mathtools}
\usepackage[capitalize,noabbrev]{cleveref}
\usepackage{colortbl}
\usepackage{subcaption}
\usepackage{bbm}
\usepackage[title]{appendix}
\usepackage{multirow}
\usepackage{siunitx}
\usepackage{microtype}
\microtypecontext{spacing=nonfrench}
\usepackage{graphicx}
\usepackage{xspace}
\usepackage{xfrac}
\usepackage{booktabs} %
\usepackage[subtle]{savetrees}
\usepackage{enumitem}
\usepackage{algorithm}
\usepackage[utf8]{inputenc} %
\usepackage[T1]{fontenc}    %
\usepackage{amsfonts}       %
\usepackage{nicefrac}       %
\usepackage{algorithmicx}
\usepackage{algpseudocodex}

\usepackage[font=small, labelfont=bf, skip=1ex]{caption}

\makeatletter
\renewcommand{\paragraph}{%
  \@startsection{paragraph}{4}%
  {\z@}{1.25ex \@plus 0ex \@minus .2ex}{-1em}%
  {\normalfont\normalsize\bfseries}%
}
\makeatother

\newcommand{\squishlist}{
 \begin{list}{$\bullet$}{ 
        \setlength{\itemsep}{0pt}
        \setlength{\parsep}{3pt}
        \setlength{\topsep}{3pt}
        \setlength{\partopsep}{0pt}
        \setlength{\leftmargin}{1.5em}
        \setlength{\labelwidth}{1em}
        \setlength{\labelsep}{0.5em} } }
\newcommand{\squishend}{
  \end{list}
}

\newcommand{\sys}{Glinthawk\xspace}

\newcommand{\figwidth}{0.95\linewidth}

\makeatletter
\newlength{\phaserulewidth}
\newcommand{\setphaserulewidth}{\setlength{\phaserulewidth}}
\newcommand{\phase}[1]{%
  \vspace{-1.25ex}
  \Statex\leavevmode\llap{\rule{\dimexpr\labelwidth+\labelsep}{\phaserulewidth}}\rule{\linewidth}{\phaserulewidth}
  \Statex\strut\textit{#1}%
  \vspace{-1.25ex}
  \Statex\leavevmode\llap{\rule{\dimexpr\labelwidth+\labelsep}{\phaserulewidth}}\rule{\linewidth}{\phaserulewidth}}
\newcommand{\firstphase}[1]{%
  \Statex\strut\textit{#1}%
  \vspace{-1.25ex}
  \Statex\leavevmode\llap{\rule{\dimexpr\labelwidth+\labelsep}{\phaserulewidth}}\rule{\linewidth}{\phaserulewidth}}
\makeatother
\setphaserulewidth{.3pt}

\usepackage{tikz}
\usetikzlibrary{calc}

\newcommand{\shadeline}[1]{%
    \begin{tikzpicture}[remember picture, overlay]
     \tikz[overlay,remember picture,baseline] \node [anchor=base] (begin) {};
        \fill[gray, opacity=0.15] ($(begin|-begin)+(-#1,+8pt)$) rectangle ($(begin|-begin)+(\dimexpr\labelwidth+\labelsep+\linewidth-#1\relax,-4pt)$);
     \end{tikzpicture}%
}

\renewcommand\alglinenumber[1]%
 {\ifodd\value{ALG@line}\relax
     \ifnum\value{ALG@line}<10
        \shadeline{3.75pt}
     \else
        \shadeline{0pt}
     \fi
  \fi
\footnotesize\arabic{ALG@line}:}

\usepackage[acronym]{glossaries}
\glsdisablehyper
\newacronym{ml}{ML}{Machine Learning}
\newacronym{llm}{LLM}{Large Language Model}
\newacronym{cnn}{CNN}{Convolutional Neural Network}
\newacronym{gemm}{GEMM}{General Matrix Multiply}
\newacronym{gemv}{GEMV}{General Matrix-Vector Multiply}
\newacronym{nn}{NN}{Neural Network}
\newacronym{mha}{MHA}{Multi-Head Attention}
\newacronym{gqa}{GQA}{Grouped-Query Attention}
\newacronym{mqa}{MHA}{Multi-Query Attention}
\newacronym{ffn}{FFN}{Feed-Forward Network}
\newacronym{flops}{FLOPS}{Floating-point Operations Per Second}
\newacronym[plural=KVCs, firstplural=Key/Value Caches (KVCs)]{kvc}{KVC}{Key/Value Cache}
\newacronym{rtt}{RTT}{Round-trip Time}
\newacronym{cit}{CIT}{Cost per Inference Throughput}
\newacronym{cptu}{CPTU}{Cost per Throughput Unit}
\newacronym{tco}{TCO}{Total Cost of Ownership}
\newacronym{tft}{TTFT}{Time to First Token}
\newacronym{tot}{TPOT}{Time per Output Token}
\newacronym{tit}{TPIT}{Time per Input Token}

\DeclareMathOperator*{\argmax}{arg\,max}
\DeclareMathOperator*{\argmin}{arg\,min}

\newcommand{\eg}{\emph{e.g.,}\xspace}

\theoremstyle{plain}

\theoremstyle{definition}

\theoremstyle{remark}

%% file: sections/0-abstract.tex
We introduce \sys, an architecture for offline Large Language Model (LLM) inference. By leveraging a two-tiered structure, \sys optimizes the utilization of the high-end accelerators (``Tier 1'') by offloading the attention mechanism to lower-end compute tier (``Tier 2''). This separation allows the memory demand of the attention, known as the key-value cache, to scale independently from the model weights, enabling larger batch sizes and more efficient accelerator usage.
Prototyped with NVIDIA T4 GPUs and standard CPU VMs, \sys improves throughput by $5.9\times$ and reduces cost of generation by $2.8\times$, compared to paged attention baselines. For long sequence lengths, it achieves $16.3\times$ throughput improvement at $2.4\times$ less cost. Our evaluation shows that this architecture can tolerate moderate network latency with minimal performance degradation, making it highly effective for latency-tolerant, throughput-focused applications such as batch processing. The prototype is publicly available at 
\url{https://github.com/microsoft/glinthawk}.

%% file: sections/1-introduction.tex
\section{Introduction}
\label{sec:intro}

\glspl{llm}, a class of deep neural networks characterized by their
underlying \emph{Transformer} architecture~\cite{vaswani2023attention},
have found remarkable success in natural language processing tasks~\cite{radford2019language}.
\glspl{llm} predict probability distributions over sequences of sub-words called ``tokens'' and operate autoregressively, generating text one token at a time.

Serving \glspl{llm} often requires top-of-the-line accelerators (e.g., GPUs) and interconnects (e.g., InfiniBand). Due to the high costs of such infrastructure, there is a strong interest in maximizing the number of \emph{tokens} processed per second, i.e., the inference throughput, to spread the cost across many tokens~\cite{MLSYS2023_c4be71ab, yu2022orca,kwon2023efficient,sheng2023flexgen,juravsky2024hydragen}.
This is particularly important for throughput-focused application~\cite{sheng2023flexgen}
such as mass document processing~\cite{pmlr-v225-goel23a}, model benchmarking~\cite{zheng2023judgingllmasajudgemtbenchchatbot}, and data cleaning~\cite{narayan2022foundationmodelswrangledata}, where the goal is to maximize sustained processing rates rather than minimizing per-request response times. Providers also offer services tailored to these use cases~\cite{openai2024batchrequests} at discounted rates compared to interactive sessions like chatbots. This work focuses on improving throughput for such throughput-focused, latency-tolerant applications.

\gls{llm} inference suffers from low utilization of computing resources, sometimes as low as 1\%~\cite{MLSYS2023_c4be71ab}, as the matrix multiplications are bottlenecked by the rate at which the model weights can be loaded into registers. A common technique to boost utilization is \emph{batching}, where tokens from multiple prompts are processed together to amortize the cost of loading model weights~\cite{kwon2023efficient,sheng2023flexgen,MLSYS2023_c4be71ab}.
However, the degree of batching is highly constrained by GPU memory capacity. This is due to the \emph{attention} mechanism---a key step in the computation graph of the Transformer architecture---whose efficient execution requires caching information about previously processed tokens.

\begin{figure}
    \centering
    \includegraphics[width=0.9\linewidth]{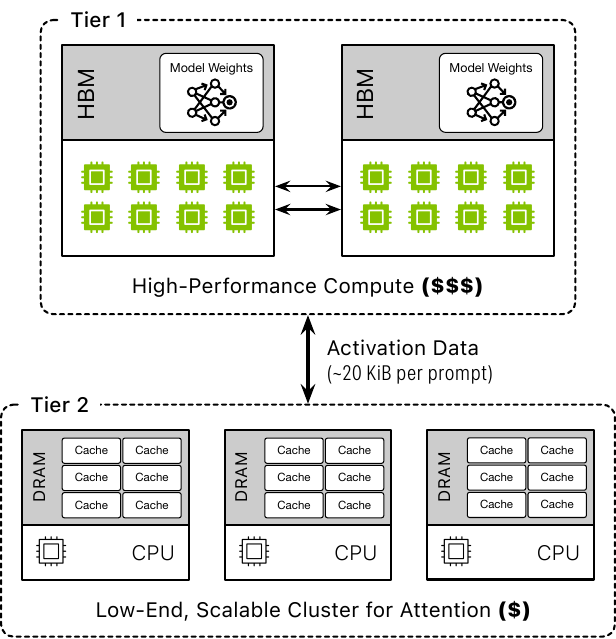}
    \caption{\sys dissects attention memory and compute to a second cluster of low-end nodes, and improves the utilization of costly GPUs working on compute-heavy operations, improving the total system throughput and reducing inference costs.}
    \label{fig:overview}
    \vspace{-0.25cm}
\end{figure}

Several efforts have been made to reduce attention memory requirements.
One line of work focuses on reducing attention's memory footprint per prompt~\cite{shazeer2019fasttransformerdecodingwritehead,ainslie2023gqa,xiao2024efficientstreaminglanguagemodels,kwon2023efficient}. Others suggest paging the attention cache in and out to secondary storage, such as host's DRAM~\cite{aminabadi2022deepspeed,kwon2023efficient} or SSDs~\cite{sheng2023flexgen}, but are severely limited by slow GPU-CPU interconnects, such as PCIe. These state-of-the-art inference engines are built with an unquestioned assumption: all computation, including attention, must be done on the high-end accelerator.

In this paper, we argue that attention is fundamentally different from other operations in the Transformer architecture: it is relatively compute-light, embarrassingly parallel, and the only stateful operation in the computation graph. Leveraging these insights, we propose a novel inference architecture that fully dissects attention from the rest of the model; storing the cache \emph{and} performing attention computation at a second set of low-end compute machinery.
This technique allows the attention subsystem to trivially scale, increasing possible batch sizes by multiple orders of magnitude. In our design, \sys,
high-performance accelerators (``Tier 1'') handle the core model computations involving
the model weights, while a set of independent, lower-end compute nodes (``Tier 2'')
manage the attention mechanism.

To demonstrate the feasibility of this approach, we implement and evaluate
one possible realization of \sys on commodity cloud infrastructure,
with NVIDIA T4 GPUs as Tier-1 accelerators, and CPU-based virtual machines
with inexpensive DRAM as the Tier-2 nodes~(\Cref{fig:overview}). These two subsystems communicate
with each other over Ethernet links.
This prototype achieves a $5.9\times$ increase in throughput and $2.8\times$ reduction in cost
compared to a traditional architecture with only the Tier-1 accelerators (details provided in \S\ref{sec:eval}).

This prototype has modest networking requirements and needs less than \SI{50}{Gbps} of inter-tier, and \SI{3}{Gbps} per node bandwidth~(\S\ref{sec:eval:net}) and can tolerate tens of milliseconds of inter-tier latency with minimal degradation in token throughput~(\S\ref{sec:eval_rtt}). It is particularly useful for processing long context prompts, outpacing a `single-tier' baseline by $16\times$ in throughput~(\S\ref{sec:eval_seqlen}). Last, we show that the aforementioned configuration is just one of many feasible realizations and highlight how \sys can be used to improve the throughput of other high-end GPUs such as NVIDIA H100s~(\S\ref{sec:eval_compute}).

The remainder of this paper is structured as follows;
in \S\ref{sec:comp_mem}, we provide a brief background
on large language models from an operational point of view, and discuss attention and non-attention operations with regards to memory bandwidth and compute load.
We analyze the inter-play of network characteristics and the cache requirements in \S\ref{sec:kvcalc}.
We discuss how we design, configure, and implement \sys in \S\ref{sec:design}, and evaluate our prototype with end-to-end benchmarks in \S\ref{sec:eval}.

We plan to release \sys as open-source software, along with the artifacts necessary to reproduce the experimental results. An anonymized version of the source code can be found at 
\url{https://github.com/microsoft/glinthawk}.

\paragraph{Throughput-oriented inference.} While interactive applications like chatbots dominate \gls{llm} usage today, where low response latency is crucial due to user engagement, there exist important throughput-oriented applications that prioritize scalability and cost over response time. These applications operate more like batch processing workloads rather than interactive request-response services. The evaluation metrics for such workloads differ fundamentally---individual response latency matters little, while overall service throughput and cost efficiency become the primary concerns. Optimizing for one class of applications often comes at the expense of the other, necessitating distinct design choices tailored to their respective demands.

This distinction provides key opportunities for designing inference infrastructure tailored to offline inference workloads. Unlike interactive inference that demand tight coupling between computation and communication---high-end accelerators connected via fast interconnects like NVLink and InfiniBand---batch workloads allow for a looser coupling.
With a greater tolerance to hardware diversity, offline inference tasks can be distributed across a mix of accelerators and general-purpose processors, leveraging hardware with different cost-performance tradeoffs. Furthermore, since throughput dominates as the key performance metric, scheduling and batching strategies can be designed to maximize hardware efficiency rather than minimize per-query response time. These insights motivate a different approach to inference infrastructure, which we explore in this paper.

%% file: sections/3-mem_comp_calculus.tex
\section{Characterizing Transformers}
\label{sec:comp_mem}

In this section, we provide a brief background on the architecture of
LLMs and Transformers, with a focus on its high-level computational
characteristics. We specifically discuss
decoder-only transformers, as they are the most common architecture for
LLMs, including GPT-3~\cite{brown2020language}, Meta's Llama~\cite{touvron2023llama1, touvron2023llama2, dubey2024llama3},
and Google's Gemma~\cite{gemmateam2024gemmaopenmodelsbased}.

To motivate our two-tiered design, we argue that transformer operations have fundamentally different resource requirements; \emph{Attention} operations can be scaled quite easily across cheap low-end nodes, while \emph{Non-attention} operations require low latency high bandwidth networking for inter-node scaling, and are best scaled with batching and high-end hardware. We will demonstrate that in the traditional (single-tier) approach, accelerators are consistently bottlenecked by attention, while much better suited to perform non-attention operations at high throughput.

\subsection{Large Language Models}
\label{sec:background:llms}

Operationally, a large language model can be
abstracted as a function $\mathcal{L}$ that, given a sequence of
\emph{tokens} (words or subwords), outputs a probability distribution
over the next possible token in the sequence. This distribution is then
sampled to produce the next token in the sequence. This process is repeated
autoregressively with the new sequence until a special token marking the
end of the sequence is sampled,
or a certain maximum length is reached. The input sequence is called the ``prompt,'' while the autoregressively generated tokens are called the ``completion.''

To avoid redundant work, some intermediate
computations for past tokens can be cached and reused when generating new tokens, commonly referred
to as the \emph{key-value (KV) cache} or the context. We use these terms interchangeably in the paper.

The Transformer architecture comprises of an embedding table, multiple transformer ``layers'' and a classifier.
Each layer has its own weights and context, receives a vector of size $D$, and produces a vector of the same size.
The embedding table and classifier map tokens to vectors and vice versa.
We can break down a Transformer block into three stages
{\small
\begin{align*}
&\mathcal{F}_1(W_k; x_k) \to qkv \\
&\mathcal{F}_2(C_k, qkv) \to C_k, x_k' \\
&\mathcal{F}_3(W_k; x_k, x_k') \to x_{k+1},
\end{align*}
}
where $x_k$ marks the activation input to layer $k$. In this notation, $\mathcal{F}_2$ is the attention mechanism that
operates on the context $C_k$ and the output from $\mathcal{F}_1$, whereas
the non-attention operations, $\mathcal{F}_{1, 3}$, deal with the
immutable model weights. This framework divides
Transformer operations into two groups, which we refer to as attention and non-attention operations hereafter.

\paragraph{Non-attention operations.} The bulk of computation in
$\mathcal{F}_{1, 3}$ consists of multiple \glspl{gemm} between model weights
and transformed inputs. Since weights are the same for
all inputs passing through a layer, their memory access cost---i.e., loading model parameters into registers for multiply-accumulate operations---can
be amortized across multiple inputs, a technique known as \emph{batching}. The memory requirements of
$\mathcal{F}_{1,3}$ remains relatively constant, with negligible increase
in the working set, as the batch size grows.

\paragraph{Attention operation.} In contrast, $\mathcal{F}_{2}$ doesn't use pre-trained weights, and only needs prompt-specific contexts and the activation input.
Attention is mainly \gls{gemv} operations between activations and their cache, and must be repeated for each prompt independently, offering no benefit from batching. In fact, batching requires more memory for prompt contexts, which often becomes a limiting factor well before other system resource limits are reached~\cite{kwon2023efficient}.

\subsection{Resource Analysis}
\label{sec:comp_mem:resources}

\input{tables/2-naming}

We focus on two key metrics: (1) number of memory load/store operations which reflects the demand
on memory bandwidth, and (2) the minimum number of floating-point operations (FLOPs) required for computation. We use the Llama2 model family~\cite{touvron2023llama2} as a running example throughout the following sections, but these arguments hold for other decoder-only Transformers as well. \Cref{table:transformer_naming} describes the symbols we will use in these discussions. 

\paragraph{Non-attention Operations:}
\label{sec:non-att-mem}
Excluding attention, and some smaller operators such as \texttt{RMSNorm} and residual connections, a Transformer layer consists of several key tensor computations, including the generation of \emph{query}, \emph{key}, and \emph{value} vectors, as well as four projections in the \gls{ffn} stage, all of which are \gls{gemm} operations. The total memory and compute requirements of these operations are depicted in \Cref{table:op_char}. For a breakdown of individual operations along with tensor sizes, refer to \Cref{table:app:op_char} in Appendix \ref{app:resource}.

\input{tables/2-mem-compt-summary}

\paragraph{Cost.}
At small batch sizes, much of the kernel time is spent on memory loads. While memory pressure is relatively constant with batch size ($B\ll D$), compute load scales linearly with it, and these operations are far more efficient at larger batches where load cost is amortized over the batch.
Eventually we cross into compute-bound territory, with no discernible benefits in increasing batch size. These batch sizes are the optimal operating point for non-attention operations.

\paragraph{Scaling.}
Compute power can be linearly scaled with parallelism, e.g., pipeline or tensor parallelism~\cite{megatron2019,liu2020deep}. \gls{gemm} operations are weakly scalable, i.e., they can be parallelized but require synchronization barriers that involve communication across nodes. The synchronization latency has to be comparable to the computation time, and with commodity networks, the gap is significant.

\paragraph{Attention Operations:}
\label{sec:att-mem}
Attention consists of a set of small, independent operations. Per each prompt in a batch, we need $H$ pairs of \gls{gemv} operations. The memory and compute load of these operators are shown in \Cref{table:op_char}. Notably, both compute and memory transfers scale linearly with batch size, as there are no common weights in attention---each prompt operates on its own dedicated set of tensors.

\paragraph{Cost.}
Attention does not become more efficient in larger batches. In fact, it would need more memory for context storage, which limits how large the batch size can grow. On the other hand, attention is far less compute-heavy than non-attention; the compute to memory load ratio for non-attention is approximately $B$, but is a constant $D/D_{kv}$ for attention.

These points have important implications. First, attention can be carried out on low-end machines with weak compute and large memory resources. For example, in \S\ref{sec:eval}, we show that CPU nodes can perform attention operations, and keep up with GPUs computing non-attention operations. Second, the cost of running attention is not dominated by compute resources, but on a balance of compute and memory. The most cost-effective approach to running attention is to find compute nodes where the cost per\gls{flops} and memory operations is low.

\paragraph{Scaling.}
Attention is \emph{embarrassingly parallel}: batches can be parallelized across both the batch dimension and the head dimension (e.g., Llama models have 32--128 heads). This means, in theory, attention can be distributed across $BH$ compute nodes without significant overhead.\footnote{Attention can also be loosely parallelized along sequence length using a distributed implementation of FlashAttention~\cite{dao2022flashattention,lin2024infinitellmefficientllmservice}.}
In contrast, such parallelization defeats the purpose of batching for \gls{gemm} operations, i.e., amortizing memory load costs. At the extreme end where each node computes one element of a \gls{gemm} operation, the memory to compute ratio is $1$, compared to $B$ when one node computes all of the matrix.

\subsection{Verdict.}
\label{sec:comp_mem_verdict}

GEMM operations are difficult to synchronize, and are, unsurprisingly, a perfect fit for high-end GPUs. They can however be made more efficient with batching. Attention has memory requirements that scale with batching, but it can be freely scaled across various nodes without losing batching benefits. Since attention is relatively low in compute, it can even be distributed across low-end nodes. We could, in theory, run attention on a swarm of tiny nodes, as long as they can process one sequence length; this parallelization will not increase computation, memory or network operations.

%% file: tables/2-naming.tex
\begin{table}[t]
\caption{Naming conventions for Transformer parameters.}
\label{table:transformer_naming}
\footnotesize
\setlength{\tabcolsep}{3pt}
\centering
\begin{tabular}{@{}r l c r l@{}}

\toprule
Symbol & Description & & Symbol & Description \\
\cmidrule{1-2}\cmidrule{4-5}
$B$ & Batch size & & $S$ & Sequence length\\
$D$ & Activation dimension & & $H$ & Number of attention heads\\
$D_{kv}$ & Key/Value dimension & & $H_{kv}$ & Number of key/value heads\\
$D_{h}$ & Hidden dimension & & & \\

\bottomrule
\vspace{-15pt}
\end{tabular}
\end{table}

%% file: tables/2-mem-compt-summary.tex
\begin{table}[t]
\caption{Memory and compute characteristics of non-attention and attention operations per layer. Smaller operations such as \texttt{RMSNorm} are omitted.
We count each multiply-accumulate as one compute operation.}
\label{table:op_char}
\footnotesize
\centering
\setlength\doublerulesep{0.6pt}

\begin{tabular}{l l l}

\toprule
Operation & Resource & Complexity \\
\midrule

\multirow{2}{*}{Non-Attention} & Compute & $\Theta(BDD_h)$ \\
\cmidrule{2-3}
& Memory Load/Store & $\Theta((B+D)D_h)$ \\

\midrule

\multirow{2}{*}{Attention} & Compute & $\Theta(BSD)$ \\
\cmidrule{2-3}
 & Memory Load/Store & $\Theta(BSD_{kv})$ \\

\bottomrule

\end{tabular}
\vspace{-12pt}
\end{table}

%% file: sections/4-kv_calculus.tex
\section{Key-Value Cache Storage}
\label{sec:kvcalc}

In \S\ref{sec:comp_mem}, we discussed how attention,
from the perspective of compute intensity and memory bandwidth requirements, scales differently from non-attention operations. Here, we dive deeper into context memory size (key-value (KV) cache footprint). In order to avoid recomputations when generating tokens autoregressively, Transformers require holding onto large context vectors until their corresponding sequence is completed. Context vectors are large enough that they rival the the size of \gls{llm} weights~\cite{kwon2023efficient}, and limit the max number of \emph{in-flight} prompts during inference.

We examine how different parallelization strategies and network performance impact the number of in-flight batches, and consequently, the maximum batch size that can be used during inference. Using a first-order model of these dynamics, we show that in traditional parallelization techniques, the maximum batch size \emph{does not} increase linearly with the number of GPUs, and network latency imposes a significant overhead on it. In contrast, \sys can scale the maximum batch size much more effectively, and can mitigate inter-tier latency by increasing the number of in-flight batches.

\subsection{Single-GPU Setup}
\label{sec:kvc_single}

As discussed in \S\ref{sec:background:llms}, Transformers benefit from storing certain intermediate data in the form of a cache. This data---the key and value vectors---is stored per each layer and token. For any layer, all cache entries produced in prior tokens will be required for processing the next token. Therefore, the cache remains `alive' until the prompt is completed. We will call a prompt `in-flight' if it is not completed and has an allocated cache, and similarly we call a batch of prompts `in-flight' if all the prompts in the batch are incomplete.

We'll compute the memory requirements for the cache. Assuming a data width of 2 bytes (i.e., half-precision floating-point~\cite{ieee7542008}), 
a singular token and layer will require $4D_{kv}$ bytes of memory~(\Cref{table:transformer_naming} describes the symbols we use in this discussion). For a Transformer with $N$ layers processing a prompt with a length of $S$ tokens before termination, the total memory required is $M \coloneq 2NSD_{kv}$. If we intend to run computation in batches of tokens from $B$ different prompts at a time, we need to store one full entry per each, i.e., $BM$. For reference, for the Llama2-70B, we need to store $M=\SI{640}{MiB}$ of data per prompt at 2048 tokens; processing a batch of size 128 requires $BM=\SI{80}{GiB}$ of memory just to store the context.

Consider a simple case where the Transformer fully resides on one accelerator. It computes layers one at a time until the last layer, generates the next set of tokens, and moves back to computing the first layer again. The GPU is never idle---always actively working on some kernel---and anytime a prompt finishes, a new one takes its place and reuses its context memory space~\cite{yu2022orca}. Only one in-flight batch is needed, and the maximum batch size will be the same as the maximum number of in-flight prompts. Supposing $C_\text{max}$ is the free memory space after allocating \gls{llm} parameters, the maximum number of in-flight prompts is $\lfloor\frac{C_\text{max}}{M}\rfloor$, same as the maximum batch size.

\subsection{Single-Tier Parallelism}
\label{sec:kvc_strawman}

If the model is too large to fit in one GPU, we have to employ a parallelization strategy to split model weights among GPUs. Below, we discuss pipeline parallelism and how it affects context dynamics.\footnote{We do not discuss parallelism strategies based on transferring weights, e.g., FSDP, Zero-3~\cite{zhao2023pytorchfsdpexperiencesscaling, rajbhandari2020zeromemoryoptimizationstraining, ren2021zerooffloaddemocratizingbillionscalemodel}, whom have substantial network requirements.}.
We discuss tensor parallelism in Appendix \ref{app:kvcalc}; in summary, tensor parallelism is sensitive to network latency, and bottlenecks batch size growth on Ethernet links. The communication overhead is negligible under NVLink and tolerable with InfiniBand, hence the popularity of tensor parallelism under strong interconnects.

\paragraph{Pipeline parallelism.}
To scale to large transformers, we can split the model layer by layer across multiple nodes; with $K$ GPUs, each node hosts $\frac{N}{K}$ layers. Each GPU takes in a batch, passes it through the layers it hosts, and sends it to the next GPU.\footnote{We focus on \gls{llm} inference, and do not consider training. Pipeline parallelism has challenges in training, due the existence of a backward pass~\cite{liu2020deep}.} While the next GPU is processing that batch, the current GPU starts working on another batch. Once a batch has passed through the last layer, it comes back to the first layer to start the next token prediction.

The key question is how many unique `in-flight' batches are needed to keep all GPUs busy. This is important, since the maximum batch size we can run is inversely proportional to this value, i.e., if we need $\textit{IF}_\textit{pp}(B)$ in-flight batches of size $B$ for full utilization, $B$ should be such that $B \le \lfloor\frac{C_\text{max}}{M \times \textit{IF}_\textit{pp}(B)}\rfloor$. The reduction in batch size, $\textit{IF}_\textit{pp}(B)$, is the toll that transmission latency takes on throughput.

Suppose, for batch size $B$, the computation latency is $t_{c}^B$ for one layer, and the communication latency of one batch from one GPU to the next is $t_{n}^B$. If communication latency is negligible compared to computations for $\frac{N}{K}$ layers (i.e., $t_{n}^{B} \ll N/K\times t_{c}^{B}$), we need $K$ in-flight batches to keep all GPUs busy. If not, we need more than 1. With $0 \ll t_{n}^B \le N/K\times t_{c}^B$, we need $2K$ unique batches; Odd-numbered batches are in transit to the next node while even-numbered batches undergo compute, and the roles are reversed when the transit and computation is over. In general, to keep all GPUs busy we need this number of batches:
{\small\begin{equation}
    \textit{IF}_\textit{pp}(B)=\left\lceil1+\frac{Kt_{n}^B}{Nt_{c}^B}\right\rceil \times K
\end{equation}}

\begin{figure}[!t]
    \centering
    \includegraphics[width=\figwidth]{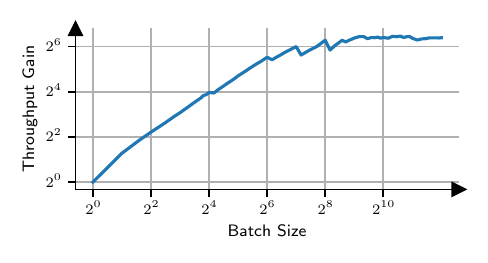}
    \caption{Throughput gain vs. batch size, for the Llama2-70B transformer running on a NVIDIA T4 GPU.}
    \label{fig:llama2-70b-profile}
\end{figure}

For instance, with the Llama2-70B transformer the computation takes $t_{c}^{1}\approx\SI{5.6}{ms}$ on an NVIDIA T4 GPU at a batch size of $B=1$. The communication involves $\SI{16}{KiB}$ of data transfer, and assuming an \gls{rtt} of $\SI{2}{ms}$ and $\SI{8}{Gbps}$ Ethernet bandwidth, takes $t_{n}^B=1+(\SI{16}{KiB}/\SI{8}{Gbps})\approx1ms$. We need $\textit{IF}_\textit{pp}(1)\approx\lceil0.022K+1\rceil$ in-flight batches for full utilization. At 10-way pipeline parallelism~($K=10$, hypothetical minimum T4 GPUs needed to evenly host the model), we need 20 in-flight batches of size $B=1$. As we have context space for $\frac{C_\text{max}}{M}=32$ prompts, $B=1$ is the maximum batch size we can use. For reference, under no network latency $\textit{IF}_\textit{pp}(\cdot)=K$ and $B=3$. The non-attention operations of this Transformer do not saturate compute until $B=256$, as observed in ~\Cref{fig:llama2-70b-profile}.

\paragraph{Fundamental limitations.}
\Cref{table:context_summary} summarize the context dynamics discussed before, along with tensor parallelism from \S\ref{app:kvcalc}. In both parallelism approaches the available context memory $C_\text{max}$ will scale linearly with the number of GPUs $K$. Unfortunately, so will the number of in-flight batches required to maintain full utilization ($\textit{IF}_\textit{pp}(B),\textit{IF}_\textit{tp}(B) \propto K$).
As such, the batch size will not improve with parallelism, and token throughput will at best scale linearly with $K$. Note that a linear increase in throughput with $K$ is not useful; We can achieve the same throughput gains by hosting $M$ parallelism sessions, with each session using $K/M$-way parallelism. 

\input{tables/4-context-summary}

Note that this limitation cannot be erased with higher-end GPUs. For example, while a single H200 GPU can host the Llama2-70B transformer from our example, it only has space for one prompt worth of context and batch size has to be $B=1$. To scale to higher batch sizes, we need to parallelize multiple H200 GPUs, which leads back to the same issue with parallelization we discussed.

\subsection{\sys: Two-Tier Parallelism}
\label{sec:kvc_sys}

Our two-tiered scheme can scale available context memory $C_\text{max}$ linearly with the number of Tier-2 nodes, with minimal impact on the number of in-flight batches. In this scheme, the context memory and attention computation is moved to a dedicated second tier of compute nodes; the first tier only works on non-attention operators, which mainly involve \gls{gemm} operations. Besides the points mentioned in \S\ref{sec:comp_mem} (attention is embarassingly parallel and computationally cheap), this architecture has another interesting benefits; \emph{attention is the only stateful computation}. Non-attention operators are stateless, i.e., the expensive GPUs computing them can be swapped, downscaled or upscaled without disrupting the entire inference. The nodes hosting attention are stateful, and difficult to alter.

\begin{figure}[!t]
    \centering
    \includegraphics[width=\figwidth]{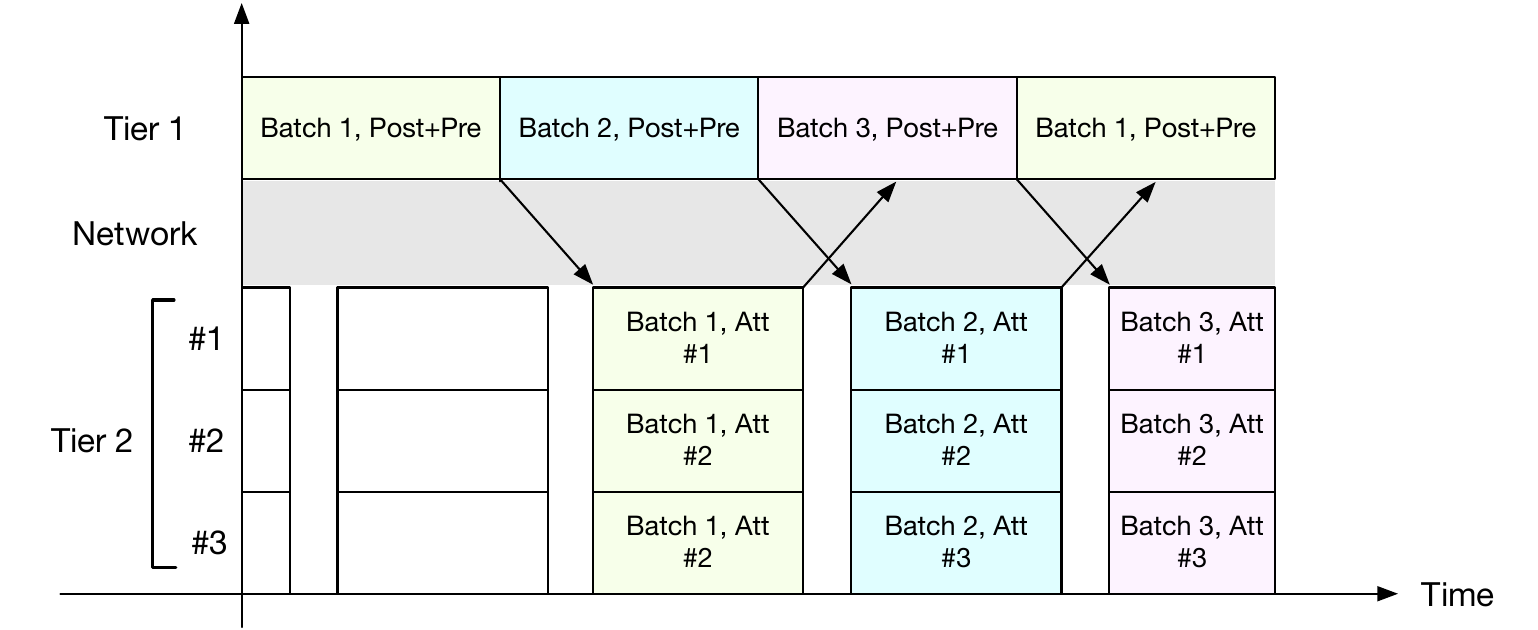}
    \caption{\sys's batching schedule. \sys hides the inter-tier transit time by utilizing multiple inflight batches.}
    \label{fig:glinthawk_schedule}
    \vspace{-10pt}
\end{figure}

\paragraph{Context dynamics.}
The \sys parallelization schedule is visualized in \Cref{fig:glinthawk_schedule}. When a Tier-1 node has finished non-attention operations on a batch of size $B\times K'$, the batch is split to $K'$ pieces of size $B$ and sent to $K'$ Tier-2 nodes. Attention operations are performed on these batches with a delay of $t_{c2,att}^{B}$, and sent back to the Tier-1 node. Suppose the overall transit time takes $t_{n}^{BK'}$. To keep Tier-1 busy in the meantime, we need other in-flight batches that undergo non-attention operations. If the Tier-1 computation latency is $t_{c1,no-att}^{BK'}$ for the full batch, we need
{\small
\begin{equation}
    \textit{IF}_\textit{gh}(B,K')=\left\lceil1+\frac{t_{c2,att}^{B}+t_{n}^{BK'}}{t_{c1,no-att}^{BK'}}\right\rceil
\end{equation}
}
in-flight batches to hide the Tier-2 total latency from Tier-1.

This overhead is practically invariant to $K'$. For small $K'$, we have network latency is negligible compared to attention computation time $t_{n}^{BK'} \ll t_{c2,att}^{B}$ and therefore $\textit{IF}_\textit{gh}(B,K')\approx\lceil 1+ t_{c2,att}^{B}/t_{c1,no-att}^{BK'} \rceil$, which does not increase with $K'$. With larger values of $K'$, we have $t_{n}^{BK'} \gg t_{c2,att}^{B}$ and $\textit{IF}_\textit{gh}(B,K')\approx\lceil1+t_{n}^{BK'}/t_{c1,no-att}^{BK'}\rceil$; Since the Tier-1 batch size $BK'$ grows large, the non-attention latency is compute-bound and grows linearly with $K'$, canceling the linear growth in transit time. Therefore, the number of in-flight batches needed, $\textit{IF}_\textit{gh}(B,K')$, does not scale with $K'$.
The optimal Tier-1 batch size $BK'$ is depicted against $K'$ in \Cref{table:llama2-tier2-overhead}, for the running example we have used thus far. The Tier-1 batch size grows super-linearly with $K'$, highlighting the scalability of this approach.

\input{tables/4-glinthawk-example}

This in-flight batch overhead is additional to whatever parallelization scheme we use to split non-attention operators in Tier 1. In this work, \sys parallelizes Tier-1 nodes with pipeline parallelism, mainly for latency resilience, but it is trivial to extend \sys to tensor parallelism.

%% file: tables/4-context-summary.tex
\begin{table*}[t]
\caption{Communication latency, minimum in-flight batches and max batch size for different parallelism schemes.}
\label{table:context_summary}
\small
\centering
\begin{tabular}{l c c c c}

\toprule
Parallelism & Context Memory & Communication latency & Minimum Inflight-Batches & Max Batch Size \\
\midrule
Single-GPU & -- & -- & 1 & $\lfloor \frac{C_{max}}{M} \rfloor$ \\
\midrule
Tensor & $C_{max} \propto K$ & $t_{n}^{B}\approx\frac{RTT}{2}+\frac{2BD}{BW}$ & $\textit{IF}_\textit{tp}(B)=\lceil1+K\frac{t_{n}^B}{t_{c,min}^B}\rceil$ & $\max_{B}\text{  if  }B \le \lfloor\frac{C_{max}}{M \times \textit{IF}_\textit{tp}(B)}\rfloor$ \\
\midrule
Pipeline & $C_{max} \propto K$ & $t_{n}^{B}\approx\frac{RTT}{2}+\frac{2BD}{BW}$ & $\textit{IF}_\textit{pp}(B)=\lceil1+\frac{Nt_{n}^B}{Kt_{c}^B}\rceil \times K$ & $\max_{B}\text{  if  }B \le \lfloor\frac{C_{max}}{M \times \textit{IF}_\textit{pp}(B)}\rfloor$ \\
\midrule
\sys & $C_{max} \propto K'\times K$ & $t_{n}^{BK'}\approx RTT+\frac{2BK'(4D+2D_{kv})}{BW}$ & $IF_{gh}(B,K')=\lceil1+\frac{t_{c2,att}^{B}+t_{n}^{BK'}}{t_{c1,no-att}^{BK'}}\rceil$ & $\max_{BK'}\text{  if  }BK' \le \lfloor\frac{C_{max}}{M \times \textit{IF}_\textit{gh}(B,K') \times \textit{IF}_\textit{pp}(BK')}\rfloor$ \\
\bottomrule

\end{tabular}
\end{table*}

%% file: tables/4-glinthawk-example.tex
\begin{table}[t]
\caption{Unique batches needed to achieve full utilization, using NVIDIA T4 GPUs as Tier-1 and AMD EPYC 7V12 (16 cores, \SI{110}{GiB} RAM) as Tier 2.}
\label{table:llama2-tier2-overhead}
\small
\centering
\begin{tabular}{l c c c c}
\toprule
\multirow{2}{*}{$K'$} & Tier-1 batch & Tier-2 batch & Overhead \\
& size $BK'$ & size $B$ & $\textit{IF}_\textit{gh}(B,K')$ \\
\midrule
1 & 72 & 72 & 3 \\
\midrule
2 & 124 & 62 & 3 \\
\midrule
4 & 252 & 63 & 3 \\
\midrule
8 & 632 & 79 & 3 \\
\midrule
16 & 1280 & 80 & 2 \\

\bottomrule
\end{tabular}
\vspace{-15pt}
\end{table}

%% file: sections/5-design.tex
\section{Design}
\label{sec:design}

So far, we have discussed how a two-tier inference architecture allows us to better utilize Tier-1 accelerators by scaling out the attention mechanism to Tier-2 nodes. In this section, we go over the practical design of this architecture (\S\ref{sec:overview}), explain how we determine configuration parameters such as batch sizes and the number of Tier 1 and 2 nodes (\S\ref{sec:config}), and discuss implementation details (\S\ref{sec:implement}).

\subsection{\sys: Architecture}
\label{sec:overview}

In designing \sys, we follow three main principles:

\paragraph{Maximizing Tier-1 Accelerator Utilization.} Achieving maximum Tier-1 utilization requires careful management of computation and communication. The system achieves this by designing the communication path to be asynchronous, ensuing that processing and transmission of batches are decoupled. This allows for continuous execution of concurrent batches without waiting for communication events, minimizing the idle time and maximizing utilization.

\paragraph{Centralized Control with Stateless Workers.} All scheduling and routing decisions are offloaded to a central dispatcher, ensuring that workers only execute predetermined operations without having to make decisions that require synchronization with other workers. Tier-1 workers only need model weights, while Tier-2 workers only retain KV cache slots, making them simple and interchangeable where possible. All the other information necessary to perform inference is encapsulated in the ``state objects'' that are passed between workers. This significantly reduces complexity at the worker level, and is a natural match for the static nature of the \gls{llm}.

\paragraph{Platform-Agnostic Execution Model.} The design abstracts computation and communication primitives for interoperability between heterogeneous hardware configurations. 

Grounded in these principles, \sys's design naturally unfolds, which we expand on as follows.

\subsubsection{Components}
Each tier comprises of a set of `worker' nodes. Worker nodes abstract inter-node communication to a \emph{Network Controller} and computations to a \emph{Compute Kernel}.

\textbf{Network Controller}'s main volume of traffic is with other worker nodes, and involves processing and routing state objects. Since nodes run at high throughput (e.g., an NVIDIA T4 GPUs as a Tier-1 nodes typically processes 10K prompts per second, or \SI{4}{Gbps} of activation data), they have been optimized to carry out these operations by avoiding memory copies and reusing previously allocated memory. The network controller induces less than \SI{100}{\micro\sec} overhead latency.

\textbf{Compute Kernel} abstracts attention and non-attention kernels the node should carry out. It takes in a batch and applies the operators it hosts to them. The compute kernel prioritizes batches in later layers/stages of the Transformer to make sure subsequent workers in the pipeline have enough work.

The compute kernel abstracts hardware choice with no relevance for other components. It could use a GPU, TPU, CPU, FPGA, etc., or even multiple machines when useful. For instance, with Mixture of Experts models that are parallelized through expert parallelism~\cite{jiang2024mixtralexperts, shazeer2017outrageouslylargeneuralnetworks,fedus2022reviewsparseexpertmodels}, the compute kernel abstracts layer experts that are hosted across several GPUs over the network. The compute kernel can even abstract a cluster of GPUs connected via NVLink switches. \sys is not concerned with how or where operations are hosted, as long as compute kernels across different nodes do not share hardware. 

\subsubsection{Inter-tier Routing}
Each Tier-1 node is in charge of a slice---one or more layers---of the Transformer (since \sys is currently based on pipeline parallelism), and has a dedicated set of Tier-2 nodes for its attention workload.
Tier-1 nodes split a ready-for-attention batch based on their context locations to `shards' and route the shards to their respective Tier-2 nodes. It later receives shards back from Tier-2 nodes, merges them back and computes the next non-attention operators.

The Tier-1 node does not necessitate that each Tier-2 node participates one shard per merge; rather, one node can participate more than its equal share or none at all. This helps to mitigate straggler effects in the merge operation.

Due to how context slots are assigned, workload is guaranteed to be equal among Tier-2 nodes. Therefore, there is no need for load balancing among them. We will discuss the details of context assignments next.

In case of a failure, a Tier-1 node can be replaced by announcing a new address, since the only retained state are the immutable model weights. However, when a Tier-2 node fails, the context memory for in-flight prompts is lost---the only option is to flush the prompts with lost context and start their inference from scratch. Our prototype currently does not have any failure recovery mechanism implemented.

\subsubsection{Dispatcher}
Finally, \sys includes a centralized controller called the \emph{Dispatcher}, logically located within the worker node responsible for processing the initial operations of the Transformer.
It serves three primary functions in managing the inference pipeline: (1) initial handshake with worker nodes, and assigning roles in the computation graph, based on its global view of the worker pool, (2) maintaining in-flight batches by organizing and tracking prompts as they move through the system~(i.e., continuous batching~\cite{yu2022orca}), (3) controlling context assignment of prompts to Tier-2 nodes.

When \sys is starting, the Dispatcher creates a set of in-flight \emph{batch state objects}. These objects store essential information related to prompts being processed, such as a global prompt identifier, a global context location identifier, the position of the latest token, generation metadata (e.g., temperature), and inter-stage communication data (e.g., activation vectors, queries, etc.).
The Dispatcher creates a finite number of batch state objects, a configurable system parameter further discussed in \S\ref{sec:config}).
Once initialized, these batches are updated and modified as prompts progress through the system.

The Dispatcher then allocates context space for each unallocated prompt in the batch state objects. It assigns a unique identifier that corresponds to a memory region within Tier-2 nodes for storing key-value cache vectors. 
This assignment ensures two key guarantees: first, no attention worker will run out of context memory while another has space available,
and second, the computation load of attention is proportionally distributed across workers. The latter is guaranteed since each Tier-2 worker is responsible for applying attention to the slots assigned to it, and the number of slots per worker is predetermined during configuration.

The Dispatcher maintains a list of queued unprocessed prompts. After creating all batch state objects, the Dispatcher fills them with prompt information from this queue, and sends them to worker nodes for processing. When a prompt reaches completion, the Dispatcher replaces the prompt in that slot with a new one from the queue. There is no need to announce the completion nor to release context assignments. These assignments will be reused for new prompts automatically.

The Dispatcher does not discriminate between input and output tokens. If there is a backlog of unprocessed prompts, and the batch size is sufficiently high, the number of tokens processed per second will be no different in `prefill' or `generation' phases. Of course, if these conditions are not met, we can create batch state objects that are filled with tokens from only one prompt. Also, if the latency of generation is important, we can speed up the prefill phase with a dedicated tier, as proposed in recent work~\cite{agrawal2023sarathiefficientllminference,patel2024splitwiseefficientgenerativellm}. 
For further discussion regarding prefill optimizations, please refer to Appendix \ref{app:prefill_opt}.

\subsection{\sys: Configuration}
\label{sec:config}

Given a set of resources in the form of eligible Tier-1 and Tier-2 nodes, we aim to maximize the number of tokens per second we can achieve, or alternatively minimize the cost of token generation. These metrics can be bottlenecked by a multitude of factors such as the maximum compute kernel processing speed, network bandwidth, maximum in-flight batches, etc. Maximizing gains is about balancing all these components in tandem.

Concretely, given a specific \gls{llm}, a \sys configuration is uniquely determined by the tuple $(K, K', B)$, where $K$ is the number of Tier-1 nodes, $K'\times K$ is the number of Tier-2 nodes, and $K'\times B$ is the running batch size in Tier 1. If we define $f(K, K', B)$ as the intended metric (e.g., throughput, cost per throughput unit, etc.) for this configuration, we aim to solve the following optimization problem:
{\small\begin{equation*}
    K^*, K'^*, B^* \leftarrow \argmax_{(K, K', B)} f(K, K', B).
\end{equation*}}
Since the optimization space is not too large, we can solve this problem with an exhaustive search over the parameter space. To compute $f(K, K', B)$, we simulate the full pipeline (pseudocode in \Cref{algo:simu_double} and \Cref{algo:simu_single} in Appendix \ref{app:simulation}).
Specifically, we model compute resources and link bandwidth as single-consumer queues that, given a batch with a known size, are occupied for a known amount of time and send the result to the next element in the pipeline. We also model the link latency, i.e., the raw minimum \gls{rtt}, as a ``delay'' element that is not occupy-able and introduces a fixed latency for each batch. 

We compute the available context memory $C_{max}$, create the maximum number of in-flight batches possible $\frac{C_{max}}{B\times K'\times M}$, and send them through the pipeline. Transient pipeline effects stabilize after 1 generated token, at which point we can measure the number of tokens generated per second.

We implement the simulation as a Discrete Event Simulation stack (300 Python LoC + 250 C++ LoC). We approximate link occupation times with an $\alpha$-$\beta$ model~\cite{HOCKNEY1994389,aashaka2023taccl,cai2021synthesizing}. For the compute resource occupation time, we run a one-time profiling of the Tier-1 and Tier-2 compute kernels at various batch sizes. The profile lasts \SI{20}{min} per accelerator type. To simulate a cluster with 80 NVIDIA T4 GPUs, 80 CPU nodes, and a maximum batch size of 4096, the full optimization takes less than 2 minutes to run. For more details, refer to Appendix \ref{app:simulation}.

\subsection{Implementation}
\label{sec:implement}

We implement \sys in C++ and CUDA~\cite{nickolls2008cuda} (16K LoC), and Python (2.4K LoC).
For Transformer operations, we use a mix of cuBLAS~\cite{nvidiaCuBLAS} routines and custom CUDA kernels for CUDA-powered nodes. For CPU-powered nodes we implement all operations with OpenMP~\cite{openmp}. Kernel computations run at FP32, while kernel results are stored in the model's native data type. We use CUDA Virtual Memory Management~\cite{nvidiaIntroducingLowLevel} to implement PagedAttention~\cite{kwon2023efficient}.

Our implementation abstracts away communication between nodes, which is handled at the host level with all data transmission carried out over userspace POSIX sockets. However, alternative approaches, such as NCCL~\cite{nvidiaNVIDIACollective} or custom transport layers, can also be implemented if needed.

We used a standard implementation of \gls{mha} with support for \gls{gqa}~\cite{ainslie2023gqa}. The implementation can be enhanced with FlashAttention~\cite{dao2022flashattention, dao2023flashattention2} and FlashInfer~\cite{githubGitHubFlashinferaiflashinfer} for higher throughput. Alternative kernel implementations can be used for all stages as long as \sys has visibility into context slot assignments, and can break layers to attention and non-attention stages.

%% file: sections/6-eval.tex
\section{Evaluation}
\label{sec:eval}

The primary goal of our proposed architecture is to improve the utilization of our most expensive and valuable resources---the Tier-1 accelerators---by satisfying attention's memory requirements by scaling up less expensive second-tier nodes. By evaluating different configurations of \sys, and comparing them against baselines described in \S\ref{sec:eval:setup}, we aim to investigate three primary questions:
(1) how much does \sys's two-tiered architecture improve throughput compared to standard pipeline parallelism? (\S\ref{sec:eval_summary}, \S\ref{sec:eval_e2e}) 
(2) How much network bandwidth do we need to sustain this throughput? (\S\ref{sec:eval:net})
(3) What are the costs associated with the proposed architecture? (\S\ref{sec:eval_summary},\S\ref{sec:eval:cost})
Moreover, using a series of microbenchmarks, we study how throughput is affected as we change inter-tier latency (\S\ref{sec:eval_rtt}) and sequence length (\S\ref{sec:eval_seqlen}) on throughput.

\subsection{Setup}
\label{sec:eval:setup}

\paragraph{Testbed.} We use a testbed comprising of NVIDIA Tesla T4 GPUs (\SI{16}{GiB} GDDR6), and AMD EPYC 7V12 CPUs (16 cores with \SI{110}{GiB} of DRAM). All nodes are equipped with \SI{8}{Gbps} networking. For various microbenchmarks, we also use other compute nodes such as an NVIDIA Quadro RTX 6000 (\SI{24}{GiB} GDDR6), NVIDIA A100 (\SI{40}{GiB} HBM2e), NVIDIA H100 (\SI{80}{GiB} HBM2e) and an NVIDIA A6000 (\SI{48}{GiB} GDDR6).

\paragraph{Metrics.} Our main focus is on two primary metrics: (1) throughput, measured as tokens per second~(\S\ref{sec:eval_e2e}), (2) cost per throughput unit, measured as U.S. Dollars per token per second~(\S\ref{sec:eval:cost}). We also report time-per-token, although token latency is a non-goal for \sys.

\paragraph{Dataset.} Since token processing time does scale with sequence length, the length of prompts does affect inference throughput. We use ShareGPT~\cite{sharegptShareGPTShare} dataset, specifically the first human message and AI response, as the prompt and completion length distribution of our experiments.

\paragraph{Models.} For these evaluations, we primarily focus on the Llama 2~\cite{touvron2023llama2,dubey2024llama3} model class families, due to their popularity and the fact many well-known open-source models such as Alpaca~\cite{alpaca}, Vicuna~\cite{vicuna2023} are based on this model class. We use the model's native FP16 weights without quantization.

\subsection{Baselines}
\label{sec:eval:baselines}

\paragraph{Single-Tier.}
This baseline utilizes \sys without a second tier, and allows for a direct analysis on how much the second tier helps. It is equivalent to standard pipeline parallelism.

\paragraph{TensorRT-LLM.}
TensorRT-LLM~\cite{githubGitHubNVIDIATensorRTLLM} is one of the leading inference engines on NVIDIA hardware, and supports pipeline and token-level parallelism for various models including Llama.

\paragraph{vLLM.}
vLLM~\cite{githubGitHubVllmprojectvllm} is a widely known inference framework that utilizes PagedAttention~\cite{kwon2023efficient} for efficient KV-cache management. 
We compare \sys against vLLM running with pipeline parallelism and/or tensor parallelism.

As discussed in Appendix \ref{app:prefill_opt}, \sys and Sarathi-serve~\cite{agrawal2023sarathiefficientllminference} are orthogonal techniques and can be merged for combined benefits. Thus, we do not consider Sarathi-serve as a baseline.

\subsection{Summary of the Results}
\label{sec:eval_summary}

\begin{figure}[!t]
    \centering
    \includegraphics[width=\figwidth]{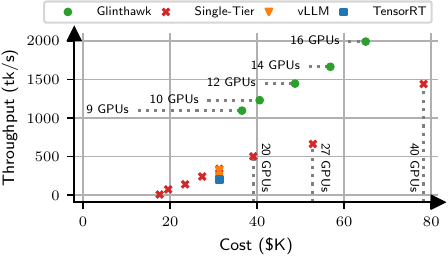}
    \caption{Inference throughput vs. setup cost for various inference engines with NVIDIA T4 GPUs as Tier 1 and AMD EPYC 7V12 16 Core CPUs as Tier 2.}
    \label{fig:motivate_example}
    \vspace{-8pt}
\end{figure}

For an end-to-end evaluation, we compare the performance of the baseline systems, as defined in \S\ref{sec:eval:baselines}, running on NVIDIA T4 GPUs to \sys supplemented with Tier-2 nodes. For a fair comparison, we report both throughput and cost for each scheme, with details regarding cost estimation in \S\ref{sec:eval:cost}. 

\Cref{fig:motivate_example} summarizes these results across various numbers of GPUs available. Despite the fact that the second tier has 45$\times$ less FLOPs than the first, \sys (12 GPUs, 36 CPU nodes, $\approx \$49K$) achieves 2.2$\times$ more throughput compared to baselines (27 GPUs, $\approx \$53K$) at less cost. As discussed in \S\ref{sec:eval:net}, the highest traffic rate for \sys nodes is less than $3Gbps$ per Tier 1 node and less than $1Gbps$ per Tier 2 node. In the next sections, we dissect each metric individually in controlled experiments.

\subsection{Throughput Analysis}
\label{sec:eval_e2e}

\begin{figure*}[!t]
    \centering
    \begin{subfigure}{0.33\linewidth}
        \includegraphics[width=\linewidth]{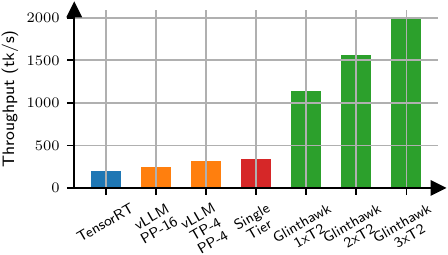}
        \caption{}
        \label{fig:e2e_thr}
    \end{subfigure}
    \begin{subfigure}{0.33\linewidth}
        \includegraphics[width=\linewidth]{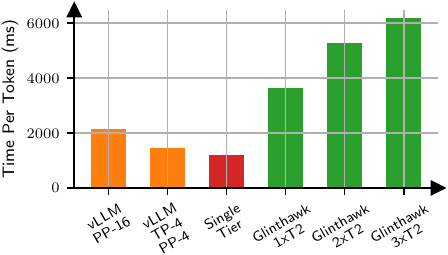}
        \caption{}
        \label{fig:e2e_tot}
    \end{subfigure}
    \begin{subfigure}{0.33\linewidth}
        \includegraphics[width=\linewidth]{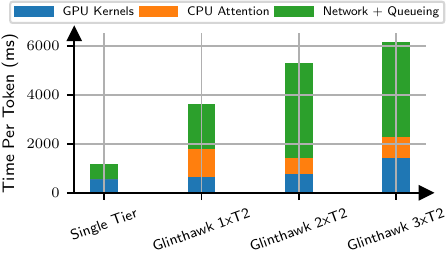}
        \caption{}
        \label{fig:e2e_tot_breakdown}
    \end{subfigure}
    \caption{\textbf{(a)} Inference throughput for various schemes using 16 NVIDIA T4 GPUs. \sys extracts more throughput from high-end Tier 1 machines compared to baselines. \textbf{(b)} Time per token across different schemes. \sys gains throughput at the cost of higher time per token, due to large batch sizes and using a second tier. \textbf{(c)} Breakdown of time per token. With more Tier 2 nodes, \sys runs at higher batch sizes and more in-flight batches, increasing compute time and queuing in GPUs.}
    \vspace{-4pt}
\end{figure*}

We compare all schemes running on a cluster of $16\times$ NVIDIA T4 GPUs, against \sys supplemented with one to three additional Tier-2 nodes per each Tier-1 node.
We aim to observe throughput benefits as we add more Tier-2 nodes.

\Cref{fig:e2e_thr} summarizes these results. Notably, our single-tier
setup achieves performance on-par or better than both vLLM (16-way pipeline parallel and 4-way pipeline parallel + 4-way tensor parallel) and TensorRT (16-way pipeline parallel). This is most likely due to the configuration optimization, discussed in \S\ref{sec:config}, that is also carried out for the single-tier baseline. Adding one CPU node per GPU boosts \sys's
throughput by $3.4\times$ over single-tier baseline. The optimal configuration for \sys, based on simulation
results, involves 3 CPU nodes per GPU, resulting in $5.9\times$ overall improvement.

\Cref{fig:e2e_tot} shows time per token (either processed or generated) for the same schemes (with the exception of TensorRT-LLM). \sys has the highest time per token (up to $5.2\times$ higher compared to the traditional pipeline parallel setup), due to several reasons, including using a higher batch size, slower attention computation on Tier 2 devices, inter-tier communication latency, and queuing.

\Cref{fig:e2e_tot_breakdown} shows the breakdown of time per token. With more Tier-2 nodes, \sys runs at higher batch sizes which increases the latency of GPU kernels; the single-tier runs at a batch size of 14, while \sys runs at batch sizes of 112, 116 and 246 with 1, 2 or 3 CPU nodes per GPU. With $1\times$ CPU, inference is bottlenecked by the rate that Tier-2 can process attention.
We can see this since when we add a second Tier-2 node, the optimizer maintains a similar batch size ($112 \to 116$) and decreases work done per CPU, but overall throughput increases.
With the addition of the third Tier-2 node, the optimizer can increase the Tier-1's batch size enough that it allows Tier-2's batch sizes to also increase, from 58 prompts ($\times2$ nodes) to 82 prompts ($\times3$ nodes) per batch.

In all these cases, more than 50\% of the time per token is spent in pending state---either being transferred over the network or waiting in a queue; according to simulations, half of the wait is due to queuing. The amount of DRAM these nodes have is nearly twice as much as needed for the minimum number of in-flight batches, and since our configuration does not optimizes for queuing, it used the maximum number of in-flight batches available. We can reduce the time per token by lowering the number of in-flight batches, if that were the goal; keeping a small queue ensures that transient slowdowns in straggling machines would not cause pipeline bubbles, as there is always some non-empty queue of batches.

We also analyze other metrics such as time per input/output token, time to first token and end to end latency in Appendix \ref{app:eval}.

\subsubsection{Simulation Fidellity}
\label{sec:eval_simulations}

\begin{figure}[!t]
    \centering
    \includegraphics[width=\figwidth]{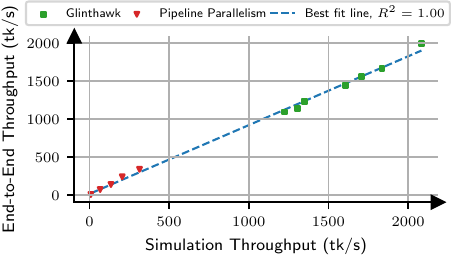}
    \caption{Predicted throughput vs. end-to-end measurements for various \sys configurations with NVIDIA T4 GPUs as Tier 1 and AMD EPYC 7V12 16 Core CPUs as Tier 2.}
    \label{fig:sim_vs_real}
\end{figure}

As discussed in \S\ref{sec:config}, we use simulations to decide the optimal configurations to run \sys at. In \Cref{fig:sim_vs_real}, we compare the empirical throughput in the end-to-end configurations tested in \S\ref{sec:eval_e2e} against the predicted throughput in simulations. As observed, the simulation matches empirical measurements with high accuracy.

\subsubsection{Hypothetical Configurations}
\label{sec:eval_compute}

\begin{figure}[!t]
    \centering
    \includegraphics[width=\figwidth]{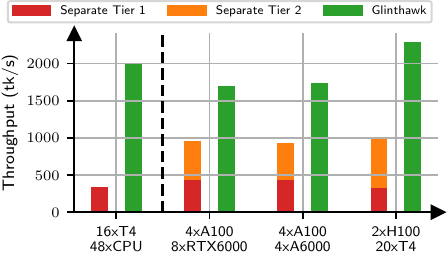}
    \caption{\sys's throughput compared to running each Tier as a separate pipeline. The T4+CPU config is from experiment data, and other configs are simulated.}
    \label{fig:alt_thr}
    \vspace{-10pt}
\end{figure}

While these experiments highlight the improvements, this is just one realization of \sys's two-tier proposal, using NVIDIA T4 GPUs and CPU nodes. \sys should be beneficial with a wide array of compute cluster pairings, e.g., using NVIDIA H100 GPUs as Tier-1 nodes and NVIDIA T4 GPUs as Tier 2. Unfortunately, we did not have access to sufficient equipment for additional hardware pairings and therefore relied on simulations. In accordance to the procedure described in \S\ref{sec:config}, we profile the performance of several high-end to medium-end GPUs, and use those real-world profiles to simulate their end-to-end performance in a first order model of the computation and networking pipeline. While these results are not end-to-end experiments, our simulations have previously matched real experiments with high fidelity, as shown in \S\ref{sec:eval_simulations}.

\Cref{fig:alt_thr} shows the predicted throughput of \sys using these clusters, compared to what these clusters could achieve as separate single-tier schemes. In all cases, the sum of these two tiers could achieve greater throughput with \sys than traditional pipeline parallelism. As covered in \S\ref{sec:comp_mem} and \S\ref{sec:kvcalc}, \sys's main strategy is to assign operations with significant differences in memory-to-compute ratio to hardware that optimally matches their resource demands.
As long as this contrast exists, \sys is effective. This disparity can be observed at many scales, whether comparing NVIDIA T4 GPUs with CPUs, or NVIDIA H100 GPUs with NVIDIA T4s.

\subsubsection{Networking Requirements}
\label{sec:eval:net}

\Cref{table:inter_bw} shows the required network bandwidth between Tier-1 and Tier-2 for end-to-end experiments and simulated configurations. These requirements are modest by commodity cloud networking standards, despite \sys's state-of-the-art token generation rate.

\input{tables/6-inter_tier_bw}

Total inter-tier traffic rate grows linearly with token throughput. Network transfers are dominated by activation data, and per each processed token, a Tier-1 node has to send activations data to Tier-2 nodes for as many layers as it hosts, i.e., $\lceil \frac{N}{K} \rceil$. Therefore, total Tier-1 egress can be approximated with
{\small
\begin{equation*}
    K \times \left\lceil \frac{N}{K} \right\rceil \times 2(2D+2D_{kv}) \times T \approx c_0 \times T,
\end{equation*}
}
where $\text{T}$ is the token throughput, and $c_0$ is a constant that depends on the \gls{llm}. Total Tier-2 egress is similar, but with smaller activation data ($2D$) leading to a smaller constant. The activation data may be amenable to compression~\cite{yuhan2024cachegen}, but that is out of the scope of this work.

\subsection{Cost Analysis}
\label{sec:eval:cost}

We evaluate \sys's throughput improvements in terms of associated costs. The fundamental aspect of our approach is the ability to meet part of the memory requirements with less expensive computing nodes, which can directly contribute to lower costs. We consider computing and memory equipment as the bulk cost of our setups,
and analyze these costs for our primary configuration (\S\ref{sec:eval_e2e}) as well as alternative configurations tested through simulation (\S\ref{sec:eval_compute}).\footnote{It is noteworthy that these costs are subject to market conditions, and this analysis may become less relevant over time.} While our prototype demonstrates the potential of \sys, we expect that reproducing this architecture in more efficient form factors could yield greater cost efficiencies over time. Custom hardware setups or optimized infrastructure could further reduce operational costs and improve performance scalability.

\begin{table}[t]
    \caption{Retail prices of the equipment used in our evaluation at the time of writing, September 2024~\cite{newegg2024newegg}.}
    \label{tab:equipment}
    \centering
    \footnotesize
    \begin{tabular}{lr}
        \toprule
        Device & Retail Price (USD) \\
        \midrule
         DDR5 Memory (\SI{128}{GiB}) & \$211 \\
         AMD EPYC 7H12 (64 cores) & \$2,078 \\
         NVIDIA T4 (\SI{16}{GiB}) & \$1,780 \\ %
         NVIDIA RTX 6000 (\SI{24}{GiB}) & \$2,280 \\
         NVIDIA A6000 (\SI{48}{GiB}) & \$4,820 \\
         NVIDIA A100 (\SI{40}{GiB}) & \$8,798 \\ %
         NVIDIA H100 (\SI{80}{GiB}) & \$30,979 \\
         \bottomrule
    \end{tabular}
\end{table}

\begin{figure}[!t]
    \centering
    \includegraphics[width=\figwidth]{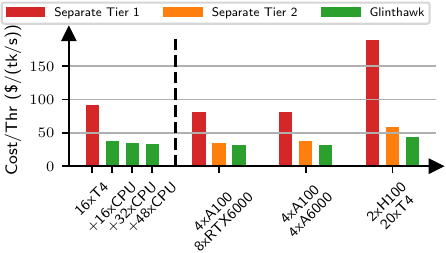}
    \vspace{0pt}
    \caption{\sys's cost per throughput unit compared to running each Tier as a separate pipeline. The T4+CPU configuration is from experiment data, and others are simulated.}
    \label{fig:alt_cost}
    \vspace{-10pt}
\end{figure}

\Cref{tab:equipment} shows the retail unit prices of the compute and memory devices used in our setup, and \Cref{fig:alt_cost} depicts what the cost per throughput unit would be across empirical and simulated schemes. \sys manages to lower the cost of inference across all schemes, particularly compared to schemes that only use high-end GPUs. Note that the cost reduction will ultimately depend on how well the tier pairs `fit', i.e., the contrast in their memory to compute. Despite this, \sys consistently achieves more throughput from the two tiers as a whole than running them separately, as observed in \Cref{fig:alt_thr}.

\subsection{Effects of Inter-Tier Latency}
\label{sec:eval_rtt}

\begin{figure*}[!t]
    \centering
    \begin{subfigure}{0.33\linewidth}
        \includegraphics[width=\linewidth]{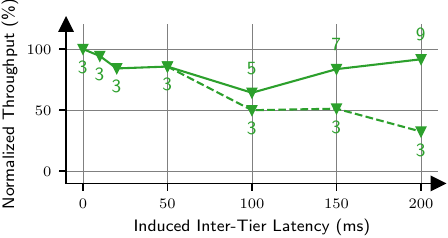}
        \caption{}
        \label{fig:rtt_thr}
    \end{subfigure}
    \begin{subfigure}{0.33\linewidth}
        \includegraphics[width=\linewidth]{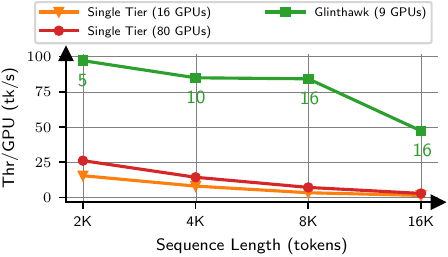}
        \caption{}
        \label{fig:seqlen_thr}
    \end{subfigure}
    \begin{subfigure}{0.33\linewidth}
        \includegraphics[width=\linewidth]{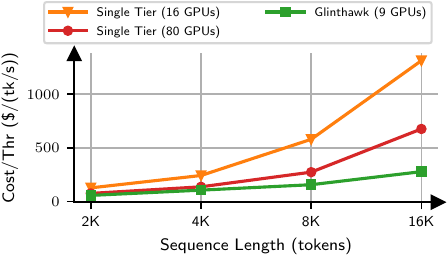}
        \caption{}
        \label{fig:seqlen_cost}
    \end{subfigure}
    \caption{\textbf{(a)} \sys's inference throughput with various inter-tier \gls{rtt} values, normalized to throughput without induced latency. The number below data points denotes the number of Tier-2 nodes per GPU. \sys mitigates inter-tier latency with more Tier-2 nodes. \textbf{(b)} Throughput per GPU and \textbf{(c)} \gls{cit} compared to a single tier setup for various sequence lengths. The number below \sys's data points denotes the number of Tier 2 nodes per GPU. As sequence length grows, KV-cache slots become more scarce, limiting batch size. This makes \sys's two-tier separation more valuable.}
\end{figure*}

To evaluate how network latency between Tier-1 and Tier-2 nodes affects \sys's performance, we induce inter-tier latency from 2ms to 200ms. \Cref{fig:rtt_thr} demonstrates the inference throughput variation at different latency values, and for different number of Tier 2 nodes per GPU. Higher latency increases the end-to-end prompt latency, which in turn increases the number of in-flight prompts, and reduce batch size. If we do not scale Tier-2 machines, throughput will drop with higher latencies. However, if we scale the Tier-2 size, we can increase in-flight batches and hide the latency from Tier-1 machines. Note that as covered in \S\ref{sec:eval:net}, the network bandwidth needed only scales with token throughput, and large inter-tier latencies do not increase required bandwidth.

For this experiment, we injected the stated latency in our code stack.
We did not perform those experiments on actual high-latency links, nor study the effects of congestion control on performance. This may be important in links with high bandwidth-delay product (BDP), and can be mended with congestion control schemes optimized for this setup.

\subsection{Scaling to Longer Sequence Lengths}
\label{sec:eval_seqlen}

Higher sequence lengths will require larger KV-cache slots and proportionally reduce the batch size the scheme can run at. However, as covered in \S\ref{sec:comp_mem} the computational load of attention scales with the context memory; regardless of sequence lengths, since the total context memory is bounded, the total computational load remains constant. In other words, Tier-2 can still keep up with Tier-1. Furthermore, \sys can scale the number of Tier-2 nodes to scale the batch size back up again. The effect can be observed in \Cref{fig:seqlen_thr,fig:seqlen_cost}. \sys outperforms the single-tier baseline both in throughput gained per GPU, and the cost per throughput unit. Note that we normalize throughput here per GPU to be able to compare across different GPU cluster sizes.

%% file: tables/6-inter_tier_bw.tex
\begin{table}[t]
\caption{Inter-tier network traffic rates (in Gbps) for several configurations and the corresponding token throughput. Since \sys only needs to send small activation data, the bandwidth requirements remain modest.}
\label{table:inter_bw}
\footnotesize
\centering
\setlength{\tabcolsep}{3pt}
\begin{tabular}{l l l c c c c c c}
\toprule
& & & & \multicolumn{2}{c}{Tier-1 Egress} & & \multicolumn{2}{c}{Tier-2 Egress} \\
\cmidrule{5-6}\cmidrule{8-9}
\multirow{8}{*}{\rotatebox[origin=c]{90}{\textit{Empirical}}}
& Tier 1 & Tier 2 & Token/sec & Per & Total & & Per & Total \\
\midrule
& $9\times$T4 & $27\times$CPU & 1096 & 2.91 & 26.2 & & 0.86 & 23.3 \\
\cmidrule{2-9}
& $16\times$T4 & $16\times$CPU & 1138 & 1.68 & 26.9 & & 1.49 & 23.9\\
\cmidrule{2-9}
& $16\times$T4 & $32\times$CPU & 1557 & 2.30 & 36.7 & & 1.02 & 32.7\\
\cmidrule{2-9}
& $16\times$T4 & $48\times$CPU & 1992 & 2.94 & 47.0 & & 0.87 & 41.8\\
\midrule\midrule
\multirow{4}{*}{\rotatebox[origin=c]{90}{\textit{Simulated}}} 
& $4\times$A100 & $8\times$RTX6000 & 1705 & 8.66 & 34.6 & & 3.85 & 30.8 \\
\cmidrule{2-9}
& $4\times$A100 & $4\times$A6000 & 1738 & 8.87 & 35.5&  & 7.88 & 31.5\\
\cmidrule{2-9}
& $2\times$H100 & $20\times$T4 & 2286 & 24.4 & 48.8 & & 2.17 & 43.4\\

\bottomrule
\end{tabular}
\end{table}

%% file: sections/8-related.tex
\section{Related Work}
\label{sec:related}

\paragraph{Efficient Kernels.}
The most direct way to improve inference is to improve the efficiency of the implemented kernels. Faster kernels leads to better utilization of GPU resources, producing a higher rate of tokens per second. On one end are techniques that offer better parallelization for models without altering the computation, such as better packing of Attention work~\cite{dao2022flashattention, dao2023flashattention2}. Orca demonstrated the effectiveness of token-level parallelism in contrast to request-level parallelism, i.e, batching tokens instead of requests~\cite{yu2022orca}. Alternatively, a line of work focuses on alterations to computation accuracy and Key Value Cache faithfulness, trading off speed for degradations in quality~\cite{krause2023smoothquant, lin2024awq}. 

\paragraph{Weight Memory.}
Another way to improve inference is through better management of weight memory locality and bandwidth needs. For high-end \glspl{llm}, model weights are commonly larger than the memory available in one GPU and need to be distributed among multiple GPUs, \eg GPT-3~\cite{brown2020language} requires 325GiB of memory to host model weights. A range of techniques exist for model distribution, \eg, model/tensor parallelism~\cite{li2023alpaserve, zheng2022alpa}, offloading~\cite{sheng2023flexgen}, pipeline parallelism~\cite{pippy2022, borzunov2022petals}.

\paragraph{Key-Value Memory.}
KV Cache memory grows linearly with batch size and sequence length, and as discussed in length in the paper, bottlenecks the maximum inference batch size.
One class of techniques focus on frugal memory management. PagedAttention demonstrated the effectiveness of allocating KV Cache as needed, instead of allocating the entire sequence memory upfront, which allowed running at higher batch sizes~\cite{kwon2023efficient}. Another line of work focuses on cases where the prompts in a batch share an initial number of tokens, and can effectively reuse the same KV Cache space~\cite{zheng2023efficiently,juravsky2024hydragen}. Some works suggest loading/unloading KV Cache to/from GPU, such as to CPU DRAM through PCIe~\cite{aminabadi2022deepspeed} or SSDs~\cite{sheng2023flexgen}. Another approach is to quantize only the KV Cache to save space~\cite{sheng2023flexgen}. These works proved effective at reducing KV cache footprint, but they cannot ultimately scale attention memory to nominal batch sizes or growing sequence lengths~\cite{googleLongContext}. Another line of work suggests aggregating GPU memories in a cluster and running distributed attention algorithms for long context requests~\cite{liu2023ringattentionblockwisetransformers, lin2024infinitellmefficientllmservice}. Concurrent to our work, Lamina suggests similar ideas of offloading attention to a second set of memory-optimized devices~\cite{chen2024efficienteconomiclargelanguage}. However, we propose \sys as a blueprint for flexible and scalable inference clusters with commodity level networking hardware. We show that our prototype is resilient to hundreds of milliseconds of latency between tiers, while Lamina needs RDMA-level network performance to maintain improvements.

%% file: sections/9-conclusion.tex
\section{Conclusion}
\label{sec:finir}

We presented \sys, a two-tiered inference architecture for large language models, where the expensive high-end accelerators handle non-attention operations, while a swarm of lower-end resources manages attention. Through extensive experiments, we demonstrated the end-to-end performance improvements of \sys, its scaling characteristics, cost-effectiveness vs. well-known baselines, and resilience to network conditions including inter-tier latency.

While our prototype has proven successful, several new directions warrant exploration. First, alternative parallelization strategies for Tier-1 nodes, such as tensor parallelism, need to be examined to better understand their implications in \sys's architecture. Second, our analysis suggests that the ideal devices for Tier-2 would be small compute nodes with enough memory and compute for one prompt. However, the hardware and network design of these nodes is as of yet unclear. We invite the community to further explore the implications of this architecture on designing the next generation of accelerators for \gls{llm} inference.

%% file: sections/appendix/resources.tex
\section{Resource Analysis}
\label{app:resource}

In \S\ref{sec:comp_mem:resources}, we discussed analyzing the computational and memory load/store charactersitics of transformer operations. \Cref{table:app:op_char} shows a detailed account of tensor sizes, memory load and compute intensity for various sub-computations in non-attention and attention operations.

\input{tables/app-mem-compt-complete}

%% file: tables/app-mem-compt-complete.tex
\begin{table*}[t]
\caption{Memory load/store and compute characteristics of non-attention and attention operations in a single Transformer layer. Smaller operations such as \texttt{RMSNorm} are omitted.
For compute, we count each multiply-accumulate operation as one.}
\label{table:app:op_char}
\small
\centering
\setlength\doublerulesep{0.8pt}

\begin{tabular}{l l l l l}

\toprule
 & Operation & Sizes & Memory Load/Store & Compute \\
\midrule

\multirow{6}{*}{\rotatebox[origin=c]{90}{Non-Attention}} & $x^T.w_{q}$ & $w_q \in \mathbb{R}^{D \times D}$, $x \in \mathbb{R}^{B \times D}$ & $2BD+D^2$ & $BD^2$ \\
 
& $x^T.w_{kv}$ & $w_{kv} \in \mathbb{R}^{D \times 2D_{kv}}$, $x \in \mathbb{R}^{B \times D}$ & $BD+2DD_{kv}+2BD_{kv}$ & $2BDD_{kv}$ \\
 
& $x^T.w_{o}$ & $w_o \in \mathbb{R}^{D \times D}$, $x \in \mathbb{R}^{B \times D}$ & $2BD+D^2$ & $BD^2$ \\
 
& $x^T.w_{1,2}$ & $w_{1,2} \in \mathbb{R}^{D \times 2D_{h}}$, $x \in \mathbb{R}^{B \times D}$ & $BD+2DD_{h}+2BD_{h}$ & $2BDD_{h}$ \\
 
& $x^T.w_{3}$ & $w_{3} \in \mathbb{R}^{D_{h} \times D}$, $x \in \mathbb{R}^{B \times D_{h}}$ & $BD_{h}+D_{h}D+BD$ & $BDD_{h}$ \\
\cmidrule{2-5}

& Total & --- & $D(2D+3D_{h}+2D_{kv})+B(8D+3D_{h}+2D_{kv})$ & $BD(2D+3D_{h}+2D_{kv})$ \\
\midrule
\midrule
\multirow{3}{*}{\rotatebox[origin=c]{90}{Attention}} & $Q_i^T.K_i$ ($i \in \{1..H\}$) & $Q_i \in \mathbb{R}^{B \times D/H}$, $K_i \in \mathbb{R}^{D/H \times S}$ & $B(D+SD_{kv}+SH)$ & $SBD$ \\
 
& $A_i^T.V_i$ ($i \in \{1..H\}$) & $A_i \in \mathbb{R}^{B \times S}$ ,$V_i \in \mathbb{R}^{D/H \times S}$ & $B(SH+SD_{kv}+D)$ & $SBD$ \\
\cmidrule{2-5}
 
& Total & --- & $2B(D+SH+SD_{kv})$ & $2SBD$ \\

\bottomrule

\end{tabular}
\end{table*}

%% file: sections/appendix/calculus.tex
\section{Key-Value Cache Storage: Continued}
\label{app:kvcalc}

In \S\ref{sec:kvc_strawman}, we discuss how pipeline parallelism affects context memory scaling and in-flight batches. Below, we extend this analysis to Tensor Parallelism.

\begin{figure}[!t]
    \centering
    \begin{subfigure}{\figwidth}
        \includegraphics[width=\linewidth]{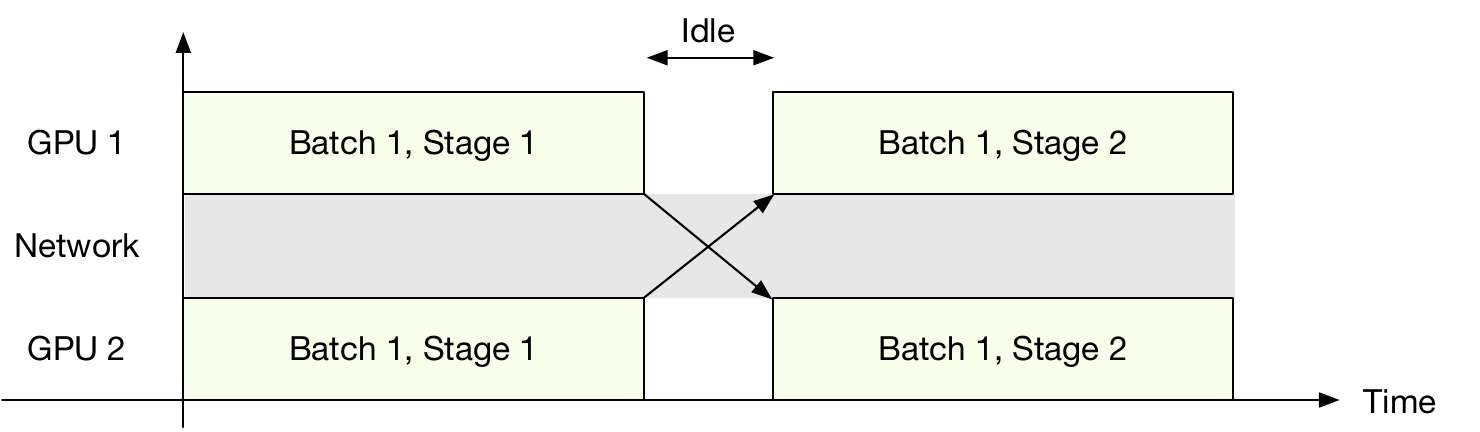}
        \caption{}
        \label{fig:tensor_sync}
        \vspace{-3pt}
    \end{subfigure}
    \begin{subfigure}{\figwidth}
        \includegraphics[width=\linewidth]{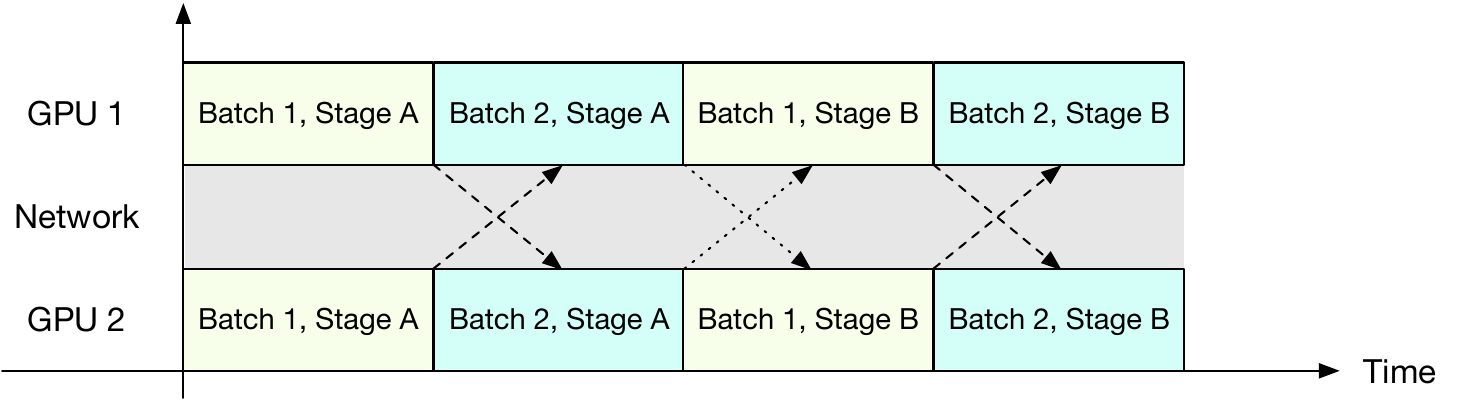}
        \caption{}
        \label{fig:tensor_async}
    \end{subfigure}
    \caption{\textbf{(a)} Synchronous vs. \textbf{(b)} asynchronous parallelism. Asynchronous parallelism has the potential to hide communication, but requires more than 1 in-flight batch.}
    \label{fig:sync_async}
\end{figure}

\paragraph{Tensor parallelism.}
In this approach, each GPU hosts a fraction of each layer, and after each computation the GPUs share their fractional results via an AllGather operator. If the GPUs run \emph{synchronously}, computation cannot continue until the AllGather operator is finished, as shown in \Cref{fig:tensor_sync}. Individual computation steps can be sub-millisecond, and this data transfer can quickly become the major bottleneck with distributed GPUs. A simple workaround is to run inference \emph{asynchronously} depicted in \Cref{fig:tensor_async}, e.g., GPUs work on stage A for batch 1, and stage A for batch 2 while batch 1 is going through AllGather.

Suppose the computation latency for the smallest stage between synchronization barriers is $t_{c,min}^{B}$ for batch size $B$ on one GPU. As tensor parallelism splits matrices to $K$ shards between $K$ GPUs, the computation would take at least $t_{c,min}^{B}/K$ per GPU node. If the communication latency of a batch $t_{n}^{B}$ is negligible compared to compute $t_{n}^{B}/t_{c,min}^{B}\approx0$, i.e., the GPUs are co-located, we only need 1 unique batch in flight at a time to keep all GPUs busy. If not, we need more than 1. With $0 \ll t_{n}^B \le t_{c,min}^B/K$, we need 2 unique batches; One batch is in transit while the other undergoes computation, and the roles are reversed when the transit and computation is over, similar to the example in \Cref{fig:sync_async}. In general, we need
\vspace{-3pt}
{\small\begin{equation}
    \textit{IF}_\textit{tp}(B):=\left\lceil1+\frac{t_{n}^B}{t_{c,min}^B/K}\right\rceil
\end{equation}}
in-flight batches to keep all GPUs busy.

For instance, with the Llama2-70B transformer the computation before the smallest synchronization barrier is $t_{c,min}^{1}\approx\SI{0.48}{ms}$ on an NVIDIA T4 GPU at a batch size of $B=1$. The communication involves $\SI{16}{KiB}$ of data transfer, and assuming an \gls{rtt} of $\SI{2}{ms}$ and $\SI{8}{Gbps}$ Ethernet bandwidth, takes $t_{n}^B=1+(\SI{16}{KiB}/\SI{8}{Gbps})\approx1ms$\footnote{The latency will be higher in practice as AllGather latency is max transit time along all nodes, and sensitive to stragglers.}. We need $\textit{IF}_\textit{tp}(1)\approx\lceil2.08K+1\rceil$ in-flight batches for full utilization. At 10-way tensor parallelism~($K=10$, hypothetical minimum T4 GPUs needed to evenly host the model), we need 22 in-flight batches of size $B=1$. As we have context space for $\frac{C_\text{max}}{M}=32$ prompts, $B=1$ is the maximum batch size we can use. For reference, under no network latency $\textit{IF}_\textit{tp}(\cdot)=1$ and $B=32$, under the specified link. The non-attention operations of this Transformer do not saturate compute until $B=256$, as observed in ~\Cref{fig:llama2-70b-profile}. The communication overhead would have been negligible under NVLink and tolerable with InfiniBand, hence the popularity of tensor parallelism under strong interconnects.

%% file: sections/appendix/simulation.tex
\section{Simulation}
\label{app:simulation}

A multitude of factors can become the bottleneck for inference in a distributed inference system, such as \sys. Mainly, these include network bandwidth for transferring activations, available memory for context, and how fast nodes are applying operation kernels. There is no straightforward way to estimate inference throughput (or latency) based on these factors.

Therefore, we run a simulation that mimics the overall pipeline of inference engines, and specifically, \sys. This simulation first estimates how many in-flight batches we can make in a given configuration---context memory needed per prompt, number of nodes in each tier, and their available memory. Then, the simulator fills all batches with dummy prompts and runs them through the Transformer layers. After several full Transformer passes, the simulator logs the time between processed/generated Tokens (TBT), and estimates the throughput based on it.

We consider the aforementioned resources in the simulation in the following manner.

\begin{itemize}
    \item To consider operation kernel times, we profile every Transformer layer operation, both separately and fused, on a sweeping range of batch sizes from 1 to 4096, logarithmically spaced at a factor of $\approx\sqrt{2}$, e.g., 1, 2, 3, 4, 6, 8, etc. For batch sizes not profiled, we use a linear interpolation of the closest batch sizes profiled, e.g., kernel durations for batch size 5 is estimated as the average of that for batch sizes 4 and 6.
    
    In simulation, each node has a priority queue of received batches. The node pulls a batch, uses the profiled kernel durations to estimate how long the processing would take, and forwards the batch to the correct node afterwards. The priority function for the queue, similar to \sys compute kernels, is based on how far along a batch is in the transformer computations, i.e., higher layers and later operations will be finished sooner. This method of prioritization is important when a node serves multiple layers or operations; since the inference engine is a pipeline, it is better for nodes to complete their part of the operations as quickly as possible to send it to the next node.

    Note that in simulation (as well as \sys), the node works on one batch at a time. 

    \item Memory is already considered when we estimate the number of available in-flight batches. An important detail arises here due to paged caching. \sys uses paged memory for context, similar to vLLM, which means \sys oversubscribes in-flight batch context memory. This is fine, because in practice most prompts do not need memory for the full sequence length, but we need an oversubscription factor here for estimating the number of in-flight batches. We have found a factor of 3$\times$ to work well.
    
    Note that due to the randomness of how long prompt generations actually are, it is possible for a node to be responsible for prompts that all need full sequence lengths. When this happens, an Out Of Memory (OOM) error is triggered, and some contexts need to be moved to DRAM/SSD/HDD. This slows down the entire pipeline and should best be avoided.
    
    The chances of OOM are astronomically low when a node hosts context for hundreds of prompts, but can be quite likely when a node only hosts context for a handful. This does happen for single-tier pipelines, and we had to reduce the factor to 2-2.5$\times$ to avoid out of memory errors.

    \item Network effects comprise of two types of delays. There is a static delay incurred due to the minimum Rount Trip Time (RTT) between nodes, and there is the transfer time of messages due to link capacity. We model these effects separately.

    For raw latency, we simply add a fixed delay equal to half of the minimum RTT---we are assuming symmetric link latencies---to batch arrival times.

    For transfer time, we model the link as a First-In-First-Out (FIFO) queue with a depletion rate equal to the link capacity. For each batch size, we compute the delay according to the batch activation memory size divided by link capacity, on top of the queueing delay.
\end{itemize}

\begin{figure}[!t]
    \centering
    \begin{subfigure}{\figwidth}
        \includegraphics[width=\linewidth]{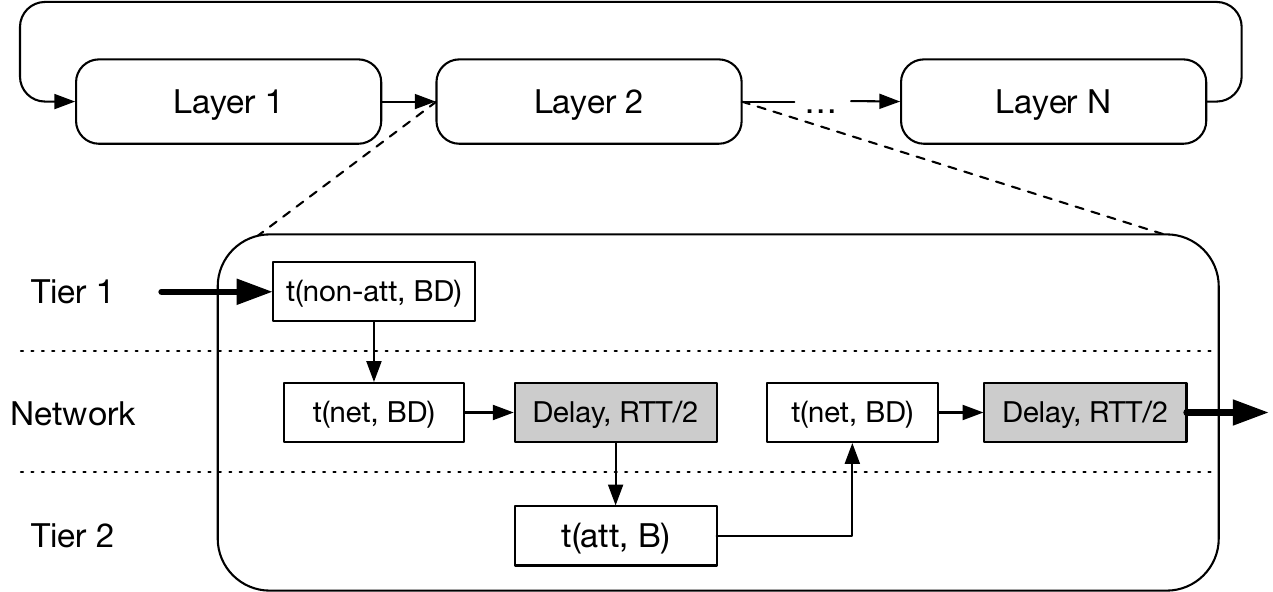}
        \caption{}
        \label{fig:des_sim_double}
    \end{subfigure}
    \begin{subfigure}{\figwidth}
        \includegraphics[width=\linewidth]{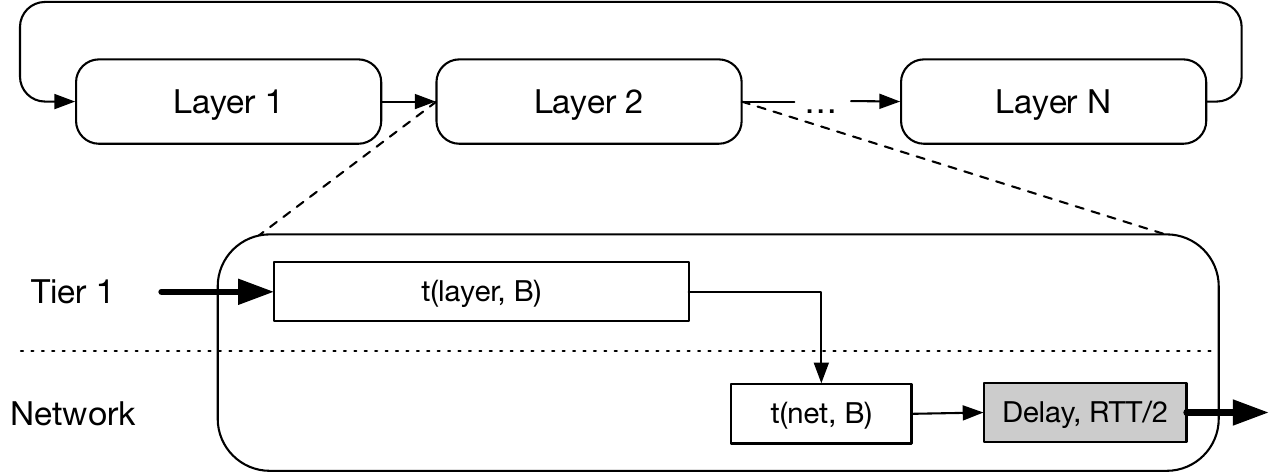}
        \caption{}
        \label{fig:des_sim_single}
    \end{subfigure}
    \caption{We simulate \textbf{(a)} \sys and \textbf{(b)} single-tier pipeline parallelism using one-time collected profiles of Tiers 1 and 2. White boxes denote queues, and gray boxes denote static additive delays.}
\end{figure}

Overall, these batches go through the queues in the order they would in \sys, which is depicted in \Cref{fig:des_sim_double}. The algorithm for this discrete event simulator is presented in \Cref{algo:simu_double}. We also simulate the single-tier engine very similarly, according to the order in \Cref{fig:des_sim_single}, and the algorithm presented in \Cref{algo:simu_single}. We allow in-flight batches to make two passes from layers 1 to N for the pipeline to stabilize. Afterwards, we log the arrival times of batch \#1 in the first layer. The difference between these arrivals is the TBT. To calculate throughput, note that in one pass we have $IF(B, K')$ in-flight batches, each holding $B$ prompts. Therefore, $B\times IF(B, K')$ tokens are processed in one pass, and we can estimate throughput with $\frac{B\times IF(B, K')}{TBT}$.

\input{algorithms/two_tier_sim}
\input{algorithms/one_tier_sim}

Overall, this simulation is highly accurate. We compare the results to end-to-end measurements in \Cref{fig:sim_vs_real} in \S\ref{sec:eval_simulations}. We do note that there are several limitations to this simulation, which we outline below:

\begin{enumerate}
    \item In calculating kernel durations and network transfers, we are assuming stable performance. In other words, this simulation does not take stragglers into account. This is not because it would be technically difficult to implement; we simply have to sample kernel times and network transfers from profiles. Rather, stragglers need to be directly resolved instead of accounting for them in performance optimization. Slow nodes need to be identified and replaced. While we leave a more nuanced study to future work, the simplest approach is to transfer context memory in phases, similar to Splitwise~\cite{patel2024splitwiseefficientgenerativellm}.
    \item We are assuming the overhead incurred by the Dispatcher, Tier Router and Network Controllers are negligible. This is true in practice, since the overhead amounts to tens of microseconds, compared to tens of milliseconds for network delays and kernel compute times.
    \item For attention kernels, we use the runtimes for when sequence length is at max. The amount of computation in the attention operation scales with sequence length, and since in practice sequence length are lower, we are overestimating how long tier-2 nodes will take to finish computations, and overestimate how many tier-2 nodes are needed to keep up with tier-1. In the future, we will amend the simulator to sample sequence lengths from real workloads for more accurate estimates.
    \item To calculate network delays, we are using the `alpha-beta' model~\cite{HOCKNEY1994389}, used in prior works concerning distributed inference and training~\cite{aashaka2023taccl,cai2021synthesizing}. In practice, congestion control algorithms incur transient delays until link bandwidth is fully utilized. This did not affect our results by much.
\end{enumerate}

%% file: algorithms/two_tier_sim.tex
\begin{algorithm*}[ht]
\caption{\sys Simulation}
\label{algo:simu_double}
\begin{algorithmic}[1]
\firstphase{Arguments}
\State Argument $N$ (number of layers).
\State Argument $K$ (number of Tier-1 nodes).
\State Argument $IF$ (number of in-flight batches).
\State Argument $E$ (number of iterations).
\State Argument $t_{net, 1\rightarrow2}$ (transit time from Tier-1 to Tier-2 due to link BW).
\State Argument $t_{net, 2\rightarrow1}$ (transit time from Tier-2 to Tier-1 due to link BW).
\State Argument $t_{net, rtt}$ (Inter-Tier RTT).
\State Argument $t_{c,att}$ (compute time for attention).
\State Argument $t_{c, non-att}$ (compute time for non-attention).

\phase{Definitions}
\State Define `stage' as iterator from 0 to $6N-1$, where each layer has 6 stages: 1) Tier-1 non-attention operations, 2) Tier-1 to Tier-2 transfer, 3) Inter-Tier RTT, 4) Tier-2 attention operations, 5) Tier-2 to Tier-1 transfer, 6) Inter-Tier RTT.
\State Define `id' as unique batch id iterator from 0 to $IF-1$.
\State Define `event' as a tuple <stage, timestamp, id>.

\phase{Initializations}
\State Initialize a latency list $larr[6]=\{t_{c, non-att}, ~t_{net, 1\rightarrow2}, ~t_{net, rtt}/2, ~t_{c,att}, ~t_{net, 2\rightarrow1}, ~t_{net, rtt}/2\}$.
\State Initialize $pipes \gets$ list of $6K$ empty event queues sorted by timestamp. Per each Tier-1 node, 6 stages comprise of 1) Tier-1 node, 2) Tier-1 to Tier-2 link, 3) Inter-Tier RTT, 4) Tier-2, 5) Tier-2 to Tier-1 link, 6) Inter-Tier RTT.
\State Initialize $stage\_to\_pipes \gets$ mapping that goes from stage number to corresponding `pipe'. This is important when some \shadeline{3.8ex}nodes host multiple layers.
\State Initialize $busy\_till\_ts \gets$ a list of $6K$ timestamp values, initially 0. Denotes until when a pipe is busy.
\State Initialize $next\_event\_ts \gets$ a list of $6K$ timestamp values, initially $\infty$. Denotes earliest time a pipe can process an event.
\State Initialize $gen\_ts \gets$ an empty list. Denotes times when batch with id 0 generated tokens.

\phase{Simulation Loop}
\For{$i$ from 0 to $IF-1$}
    \State $pipes[0].push(<0, 0, i>)$ \Comment{Add in-flight batches as events in the first worker}
\EndFor
\State $next\_event\_ts[0] \gets 0$ \Comment{First queue starts events at t=0.}
\While{$size(gen\_times) < E$}
    \State $w \gets \argmin_{0\le i<6K} next\_event\_ts[i]$ \Comment{Get worker with nearest event}
    \State $<stage, ~timestamp, ~id> \gets pipes[w].pop(.timestamp < next\_event\_ts[w])$
    \Statex \Comment{From all batches in worker \textbf{before} $next\_event\_ts[w]$, pop the batch with 1) earliest stage and 2) earliest timestamp}
    \State $next\_event\_ts[w] \gets max(busy\_till\_ts[w], pipes[w].min().timestamp)$ \Comment{Update next event times}

    \State $delay \gets larr[stage \mod 6]$ \Comment{Calculate stage delay.}
    \If {$stage \mod 3 ~!=~ 2$}
        \State $finish\_ts \gets \max(busy\_till\_ts[w],~timestamp) + delay$ \Comment{Calculate stage finish time}
        \State $busy\_till\_ts[w] \gets finish\_ts$ \Comment{Only one event is processed at a time in nodes and links}
    \Else
        \State $finish\_ts \gets timestamp + delay$ \Comment{RTT pipes are static delays, and not really queues}
    \EndIf
    \State $stage \gets (stage + 1) \mod 6N$ \Comment{Advance event to next stage}
    \State $nw \gets stage\_to\_pipes[stage]$ \Comment{Find next worker}
    \State $pipes[nw].push(<stage, ~finish\_ts, ~id>)$ \Comment{Push event to next worker}
    \State $next\_event\_ts[nw] \gets max(busy\_till\_ts[nw], pipes[nw].min().timestamp)$ \Comment{Update next event times}
    
     \If{$stage ~==~ 0$ and $batch-id ~==~ 0$} \Comment{The next event is generating a token for batch-id 0}
        \State $gen\_ts.append(timestamp)$ \Comment{Log generation time}
    \EndIf
\EndWhile
\State \Return $gen\_ts$ \Comment{Return token generation times for batch with id 0}
\end{algorithmic}
\end{algorithm*}

%% file: algorithms/one_tier_sim.tex
\begin{algorithm*}[ht]
\caption{Single-Tier Simulation}
\label{algo:simu_single}
\begin{algorithmic}[1]
\firstphase{Arguments}
\State Argument $N$ (number of layers).
\State Argument $K$ (number of Tier-1 nodes).
\State Argument $IF$ (number of in-flight batches).
\State Argument $E$ (number of iterations).
\State Argument $t_{net, link}$ (transit time from one Tier-1 node to another due to link BW).
\State Argument $t_{net, rtt}$ (Inter-Node RTT).
\State Argument $t_{c}$ (compute time).

\phase{Definitions}
\State Define `stage' as iterator from 0 to $3N-1$, where each layer has 3 stages: 1) Tier-1 non-attention and attention, 2) Tier-1 to Tier-1 transfer, 3) Inter-Node RTT.
\State Define `batch-id' as unique iterator from 0 to $IF-1$.
\State Define `event' as a tuple <stage, timestamp, batch-id>.

\phase{Initializations}
\State Initialize a latency list $larr[3]=\{t_{c}, t_{net, link}, t_{net, rtt}/2\}$.
\State Initialize $pipes \gets$ list of $3K$ empty event queues sorted by timestamp. Per each Tier-1 node, 3 stages comprise of 1) Tier-1 node, 2) Tier-1 to Tier-1 link, 3) Inter-Node RTT.
\State Initialize $stage\_to\_pipes \gets$ mapping that goes from stage number to corresponding `pipe'. This is important when some \shadeline{3.8ex}nodes host multiple layers.
\State Initialize $busy\_till\_ts \gets$ a list of $3K$ timestamp values, initially 0. Denotes until when a pipe is busy.
\State Initialize $next\_event\_ts \gets$ a list of $3K$ timestamp values, initially $\infty$. Denotes earliest time a pipe can process an event.
\State Initialize $gen\_ts \gets$ an empty list. Denotes times when batch with id 0 generated tokens.

\phase{Simulation Loop}
\For{$i$ from 0 to $IF-1$}
    \State $pipes[0].push(<0, 0, i>)$ \Comment{Add in-flight batches as events in the first worker}
\EndFor
\State $next\_event\_ts[0] \gets 0$ \Comment{First queue starts events at t=0.}
\While{$size(gen\_times) < E$}
    \State $w \gets \argmin_{0\le i<3K} next\_event\_ts[i]$ \Comment{Get worker with nearest event}
    \State $<stage, ~timestamp, ~id> \gets pipes[w].pop(.timestamp < next\_event\_ts[w])$
    \Statex \Comment{From all batches in worker \textbf{before} $next\_event\_ts[w]$, pop the batch with 1) earliest stage and 2) earliest timestamp}
    \State $next\_event\_ts[w] \gets max(busy\_till\_ts[w], pipes[w].min().timestamp)$ \Comment{Update next event times}

    \State $delay \gets larr[stage \mod 3]$ \Comment{Calculate stage delay.}
    \If {$stage \mod 3 ~!=~ 2$}
        \State $finish\_ts \gets \max(busy\_till\_ts[w],~timestamp) + delay$ \Comment{Calculate stage finish time}
        \State $busy\_till\_ts[w] \gets finish\_ts$ \Comment{Only one event is processed at a time in nodes and links}
    \Else
        \State $finish\_ts \gets timestamp + delay$ \Comment{RTT pipes are static delays, and not really queues}
    \EndIf
    \State $stage \gets (stage + 1) \mod 3N$ \Comment{Advance event to next stage}
    \State $nw \gets stage\_to\_pipes[stage]$ \Comment{Find next worker}
    \State $pipes[nw].push(<stage, ~finish\_ts, ~id>)$ \Comment{Push event to next worker}
    \State $next\_event\_ts[nw] \gets max(busy\_till\_ts[nw], pipes[nw].min().timestamp)$ \Comment{Update next event times}
    
     \If{$stage ~==~ 0$ and $batch-id ~==~ 0$} \Comment{The next event is generating a token for batch-id 0}
        \State $gen\_ts.append(timestamp)$ \Comment{Log generation time}
    \EndIf
\EndWhile
\State \Return $gen\_ts$ \Comment{Return token generation times for batch with id 0}
\end{algorithmic}
\end{algorithm*}

%% file: sections/appendix/prefill.tex
\section{Prefill and Decode Phases}
\label{app:prefill_opt}

LLM inference can be broken down to two phases. When we are processing input tokens, i.e., the prefill phase, we can possibly batch and process all tokens simultaneously in one full pass over the transformer. When we are producing output tokens, i.e., the decode phase, we generate a single token at each transformer pass, and have to redo these passes as many times as needed before the generation finishes.

Prefill tokens can be batched without worrying about the context size on the GPU. This is because all tokens belong to one prompt, and we need only one prompt's worth of context allocated. In decode, however, batching can only be done with other prompts. Therefore, batching in decode requires extensive context memory.

Inference engines usually separare prefill from decode. By running prefill at near optimal batch sizes, the Time to First Token (TTFT) and amortized inference throughput can be greatly improved. Sarathi-serve~\cite{agrawal2023sarathiefficientllminference} is an example of such engines, that merges prefill and decode batches to benefit from prefill batching without incurring head of line blocking by prefill batches. If there is no prefill phase left, batches are run at full decode up to the batch size allowed by context memory.

With minute changes, \sys can be merged with Sarathi-serve to have batches that have mixes of prefill and decode phases. As explained in \S\ref{sec:overview}, the Dispatcher decides which tokens are placed in empty slots of batches. The Dispatcher does observe whether a prompt is in prefill or decode, but currently ignores this fact. We can change the Dispatcher's strategy to match Sarathi-serve, by allocating a fraction of the batch for prefill---if prefill prompts are available---and keeping the rest for decode. The attention kernels already support such mixed batches.

Merging \sys and Sarathi-serve will improve throughput when \sys is not running under full load, i.e., there are not enough prompts to fill batches. If there are enough prompts, \sys can increase batch sizes to near optimal values. For example, in the prototype in \S\ref{sec:eval_e2e}, \sys is running at a batch size of 246, which is nearly throughput optimal according to \Cref{fig:llama2-70b-profile}. The throughput benefit of prefill is that it allows batching without needing extensive context memory, but \sys's two-tier system already provides enough context memory for full batch sizes. Of course, even under full load, merging \sys and Sarathi-serve will improve TTFT.

Splitwise~\cite{patel2024splitwiseefficientgenerativellm} is another inference engine that builds on the same prefill separation concept. Splitwise runs prefill on a separate high-end GPU, and sends the resulting context memory (key value caches of input tokens) over \SI{200}{Gbps} InfiniBand links to a different set of GPUs for decode. Splitwise uses more expensive machines (e.g., NVIDIA H100 GPUs) for prefill, where large batch sizes can be ran with low context memory, and runs the memory-hungry decode phase on less expensive machines (e.g., NVIDIA A100 GPUs) with less costly memory. This is similar to \sys's basic principle. However, \sys does not need transferring large amounts of context ($\approx$\SI{300}{MiB} per prompt) over the network, because \sys separates layer operations across machines instead of prefill/decode phases. This also allows \sys to use far cheaper hardware for the second tier (compared to NVIDIA A100 GPUs), such as CPU instances.

%% file: sections/appendix/eval.tex
\section{Evaluation: Continued}
\label{app:eval}

\begin{figure}[!t]
    \centering
    \includegraphics[width=\linewidth]{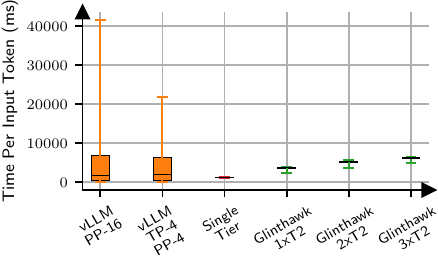}
    \caption{Box plots of Time per input token. Black lines denote median, box sides denote the first and third quantiles and whiskers denote 5th and 95th percentiles.}
    \label{fig:e2e_tpit}
\end{figure}

\begin{figure}[!t]
    \centering
    \includegraphics[width=\linewidth]{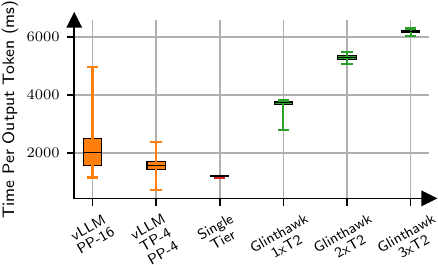}
    \caption{Box plots of Time per output token. Black lines denote median, box sides denote the first and third quantiles and whiskers denote 5th and 95th percentiles.}
    \label{fig:e2e_tpot}
\end{figure}

\begin{figure}[!t]
    \centering
    \includegraphics[width=\linewidth]{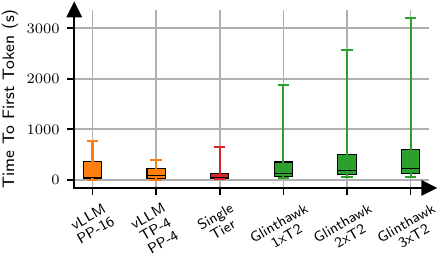}
    \caption{Box plots of Time to first token. Black lines denote median, box sides denote the first and third quantiles and whiskers denote 5th and 95th percentiles.}
    \label{fig:e2e_ttft}
\end{figure}

\begin{figure}[!t]
    \centering
    \includegraphics[width=\linewidth]{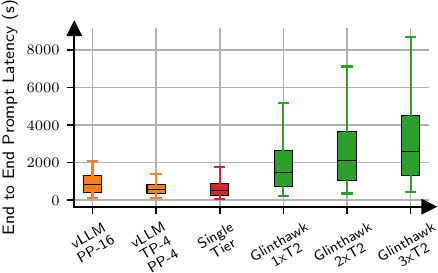}
    \caption{Box plots of End to end prompt latency. Black lines denote median, box sides denote the first and third quantiles and whiskers denote 5th and 95th percentiles.}
    \label{fig:e2e_first_2_last}
\end{figure}

In \S\ref{sec:eval_e2e} we analyzed throughput and time per token for various schemes. While \sys does not aim to optimize for latency, we plot various latencies in \Cref{fig:e2e_tpit,fig:e2e_tpot,fig:e2e_ttft,fig:e2e_first_2_last}.

\Cref{fig:e2e_tpit} shows the time per input token (TPIT) for various schemes. It should be noted that while TPIT is higher for vLLM compared to other baselines, vLLM is actually processing input tokens more quickly due to prefill. However, since all prefill tokens finish processing at the same time (or in stages), TPIT is inflated for this scheme. 

\Cref{fig:e2e_tpot} shows time per output token (TPOT) for various schemes. It is noteworthy that \sys has stable TPOT and TPIT values since, first, the overall batching and processing schedule for \sys is tightly optimized, and second, we do not use prefill separated processing.

\Cref{fig:e2e_ttft} shows time to first token (TTFT), which will depend on the dataset used, and the variance observed in the distributions is due to this fact.  

\Cref{fig:e2e_first_2_last} shows the full processing latency of prompts, from start of processing to completing the generation phase.

For clarity, it should be noted that \Cref{fig:e2e_ttft,fig:e2e_first_2_last} do not include queuing latency induced before prompts begin processing. While this is a natural metric to observe for throughput oriented inference engines, it would require running a benchmark with an equal number of prompts sampled from the dataset, and waiting until all have been flushed. The $10\times$ throughput difference between \sys and some baselines means these benchmarks will take several weeks to finish for some of them, and were not practical to consider with our resources.